\documentclass[journal]{IEEEtran}
\usepackage{cite}
\ifCLASSINFOpdf
\else
\fi
\usepackage[cmex10]{amsmath}
\usepackage{algorithmic}
\usepackage{url}
\usepackage{times}
\usepackage{helvet}
\usepackage{courier}
\usepackage{url}
\usepackage{amsmath,amssymb,amsthm}

\usepackage{caption}
\usepackage{subfigure} 
\usepackage{color}

\usepackage{algorithm}
\usepackage{algorithmic}

\usepackage{graphicx}

\graphicspath{{images/}} % Directory in which figures are stored

\makeatletter

\newcommand{\Rmnum}[1]{\expandafter\@slowromancap\romannumeral #1@}
\makeatother

\newtheorem{assumption}{Assumption}
\newtheorem{theorem}{Theorem}
\newtheorem{corollary}{Corollary}
\newtheorem{prop}{Proposition}
\newtheorem{lemma}{Lemma}
\newtheorem{remark}{Remark}

\def\k{\kappa}

\def\hx{\hat{x}}
\def\hy{\hat{y}}
\def\hu{\hat{u}}
\def\tx{\tilde{x}}
\def\ty{\tilde{y}}
\def\tu{\tilde{u}}
\def\tz{\tilde{z}}
\def\CE{\mathbb{E}}
\def\R{{\mathbb R}}
\def\tkappa{\tilde{\kappa}}

\begin{document}
%
% paper title
% can use linebreaks \\ within to get better formatting as desired
% Do not put math or special symbols in the title.
\title{Stochastic Variance-Reduced ADMM}
%
%
% author names and IEEE memberships
% note positions of commas and nonbreaking spaces ( ~ ) LaTeX will not break
% a structure at a ~ so this keeps an author's name from being broken across
% two lines.
% use \thanks{} to gain access to the first footnote area
% a separate \thanks must be used for each paragraph as LaTeX2e's \thanks
% was not built to handle multiple paragraphs
%

\author{Shuai Zheng and 
        James T. Kwok
		  %,~\IEEEmembership{Member,~IEEE,}
\thanks{S. Zheng and J. T. Kwok are with the Department
of Computer Science and Engineering, Hong Kong University of Science  and Technology, Hong Kong. E-mail: {szhengac, jamesk}@cse.ust.hk.}% <-this % stops a space
%\thanks{Manuscript received April 19, 2005; revised December 27, 2012.}
}

% note the % following the last \IEEEmembership and also \thanks - 
% these prevent an unwanted space from occurring between the last author name
% and the end of the author line. i.e., if you had this:
% 
% \author{....lastname \thanks{...} \thanks{...} }
%                     ^------------^------------^----Do not want these spaces!
%
% a space would be appended to the last name and could cause every name on that
% line to be shifted left slightly. This is one of those "LaTeX things". For
% instance, "\textbf{A} \textbf{B}" will typeset as "A B" not "AB". To get
% "AB" then you have to do: "\textbf{A}\textbf{B}"
% \thanks is no different in this regard, so shield the last } of each \thanks
% that ends a line with a % and do not let a space in before the next \thanks.
% Spaces after \IEEEmembership other than the last one are OK (and needed) as
% you are supposed to have spaces between the names. For what it is worth,
% this is a minor point as most people would not even notice if the said evil
% space somehow managed to creep in.

% The paper headers
\markboth{}%
{Zheng and Kwok, Stochastic Variance-Reduced ADMM}
% The only time the second header will appear is for the odd numbered pages
% after the title page when using the twoside option.
% 
% *** Note that you probably will NOT want to include the author's ***
% *** name in the headers of peer review papers.                   ***
% You can use \ifCLASSOPTIONpeerreview for conditional compilation here if
% you desire.

% If you want to put a publisher's ID mark on the page you can do it like
% this:
%\IEEEpubid{0000--0000/00\$00.00~\copyright~2012 IEEE}
% Remember, if you use this you must call \IEEEpubidadjcol in the second
% column for its text to clear the IEEEpubid mark.

% use for special paper notices
%\IEEEspecialpapernotice{(Invited Paper)}

% make the title area
\maketitle

% As a general rule, do not put math, special symbols or citations
% in the abstract or keywords.
\begin{abstract}
The alternating direction method of multipliers (ADMM) is
a powerful optimization solver 
in machine learning.
Recently, stochastic ADMM has been integrated with variance reduction methods for stochastic
gradient, leading to SAG-ADMM and SDCA-ADMM that have fast convergence rates
and low iteration complexities.
However, their space requirements can still be high. 
In this paper, we propose an integration of 
ADMM with
the method of stochastic variance reduced gradient (SVRG).
Unlike another recent integration attempt called
SCAS-ADMM,
the proposed algorithm retains the fast convergence benefits of 
SAG-ADMM and SDCA-ADMM,
but is more advantageous in that 
its storage requirement
is very low,  even independent of the sample size $n$. We also extend the proposed method for nonconvex problems, and obtain a convergence rate of $O(1/T)$.
Experimental results demonstrate that it is as fast as SAG-ADMM and SDCA-ADMM, much faster than
SCAS-ADMM, and can be used on much bigger data sets.
\end{abstract}

% Note that keywords are not normally used for peerreview papers.

% For peer review papers, you can put extra information on the cover
% page as needed:
% \ifCLASSOPTIONpeerreview
% \begin{center} \bfseries EDICS Category: 3-BBND \end{center}
% \fi
%
% For peerreview papers, this IEEEtran command inserts a page break and
% creates the second title. It will be ignored for other modes.
\IEEEpeerreviewmaketitle

%%%%%%%%%%%%%%%%%%%%%%%%%%%%%%%%%%%%%%%%%%%%%%%%%%%%%%%%%%%%%%%%%%%%

\section{Introduction}

In this big data era, tons of information are generated every day.
Thus, efficient optimization tools are needed to solve the  resultant
large-scale machine learning problems. 
In particular, the well-known stochastic gradient descent (SGD) \cite{bottou-04} and 
its variants \cite{parikh2014proximal} have drawn a lot of interest. 
Instead of visiting all the training samples in each iteration, 
the gradient is computed by using one sample or a small mini-batch
of samples. The per-iteration complexity is then reduced from $O(n)$, where $n$ is the number
of 
training samples, to $O(1)$.
Despite its scalability,
the
stochastic gradient is much noisier than the batch gradient. Thus, 
the stepsize has to be decreased gradually as stochastic learning
proceeds,
leading to slower
convergence.

Recently, a number of fast algorithms have been developed that 
try to reduce the variance of stochastic gradients
\cite{defazio-14,johnson2013accelerating,roux2012stochastic,shalev2013stochastic}.
With the variance reduced,
a larger constant stepsize can be used. Consequently, much faster convergence, 
even matching that of its batch counterpart, is attained.
A prominent example 
is the 
stochastic average gradient (SAG)
\cite{roux2012stochastic}, which 
reuses the old stochastic gradients computed
in previous iterations. A related method is
stochastic dual coordinate ascent (SDCA) \cite{shalev2013stochastic}, which 
performs stochastic coordinate ascent
on the dual.
However, a caveat of SAG is that storing the old gradients takes $O(nd)$ space, where $d$ is the 
dimensionality
of the model parameter.
Similarly, SDCA requires storage of the dual variables, which scales as $O(n)$. 
Thus, they can be expensive in applications with large $n$ (big sample size) and/or large $d$ (high dimensionality).

Moreover, many machine learning
problems, such as graph-guided fused lasso and overlapping group lasso,
are too complicated for SGD-based methods.
The alternating direction method of multipliers (ADMM) has been 
recently
advocated as an efficient optimization tool
for a wider variety of models
\cite{boyd2011distributed}.
Stochastic ADMM extensions have also been proposed
\cite{ouyang2013stochastic,suzuki2013dual,Wang2012}, though they only have suboptimal
convergence rates. Recently, researchers have borrowed variance reduction techniques 
into ADMM. 
The resultant algorithms, SAG-ADMM 
\cite{zhong2014fast}
and SDCA-ADMM
\cite{suzuki2014stochastic},
have fast convergence
rate as batch
ADMM but are much more scalable.
The downside is that they
also inherit the drawbacks
of SAG and SDCA.
In particular, 
SAG-ADMM and SDCA-ADMM
require $O(nd)$ and $O(n)$ space, respectively, to
store the past gradients and weights or dual variables.
This can be problematic 
in large multitask
learning, where the space complexities is scaled 
by $N$,
the number of tasks.
For example, in one of our 
multitask learning 
experiments, SAG-ADMM needs
$38.2$TB for storing the weights, and SDCA-ADMM needs $9.6$GB for the dual variables. 

To alleviate this problem, one can 
integrate ADMM with another popular variance reduction method, namely, stochastic variance reduced gradient (SVRG) \cite{johnson2013accelerating}.
In particular, SVRG is advantageous
in that no
extra
space for the intermediate gradients or dual variables
is needed.
However, this integration is not straightforward. A recent initial attempt is made in 
\cite{zhao2015scalable}.
Essentially, their SCAS-ADMM algorithm uses SVRG as an inexact stochastic solver for one of
the ADMM subproblems. The other ADMM variables are not updated until that subproblem has been
approximately solved. Analogous to the difference between Jacobi iteration and Gauss-Seidel
iteration, this slows down convergence. Indeed, 
on strongly convex problems, SCAS-ADMM only has 
sublinear convergence while SDCA-ADMM has a linear rate. On general convex problems, 
SCAS-ADMM requires the stepsize to be gradually reduced.
This defeats the original purpose of using SVRG-based algorithms,
which aim at using a larger,
constant learning rate to achieve fast  convergence \cite{johnson2013accelerating}. 

Besides, in spite of successful applications of ADMM to convex problems, the theoretical properties for nonconvex ADMM is not well understood and have been established
very recently. In general, ADMM may fail to converge due to nonconvexity. However, it is found that ADMM has presented very good performance on  many nonconvex problems. Indeed, the successful nonconvex applications includes matrix completion \cite{shen2014augmented},  tensor factorization \cite{liavas2015parallel} and robust tensor PCA \cite{jiang2016structured}. Recently, Hong et al. \cite{hong2016convergence} studied the convergence of the ADMM for solving
certain nonconvex consensus and sharing problems, and showed that the generated sequence will converge to a stationary point, as well as a convergence rate of $O(1/T)$ for consensus problems, which underlines the feasibility of  ADMM applications in nonconvex settings. Li and Pong \cite{li2015global} studied the convergence of ADMM for some special nonconvex composite models, and demonstrated that a stationary
point of the nonconvex problem is guaranteed if the penalty parameter is chosen
sufficiently large and the sequence generated has a cluster point. The Bregman multi-block ADMM, an extension of  classic ADMM, has also been studied for a large family of nonconvex functions in \cite{wang2015convergence}. Very recently, Wang et al. \cite{wang2015global} also obtained convergence guarantee of ADMM on minimizing a class of nonsmooth nonconvex functions. The developed theoretical analysis includes various nonconvex functions such as piecewise linear function, the $\ell_q$ quasi-norm for $q \in (0, 1)$ and Schatten-q quasi-norm ($0 < q < 1$), as well as the indicator functions
of compact smooth manifolds.  However, there is no known study for stochastic ADMM on nonconvex problems.

In this paper, we propose a tighter integration of
SVRG and ADMM
with a constant learning rate.
The per-iteration computational cost 
of the resultant SVRG-ADMM algorithm
is as low as existing stochastic ADMM methods, but yet
it admits fast linear  convergence
on strongly convex problems.
Among existing stochastic ADMM algorithms, a similar linear convergence result is only proved
in SDCA-ADMM for a special ADMM setting.
Besides,
it is well-known that the penalty parameter in ADMM can significantly affect convergence
\cite{nishihara2015general}.
While its effect on the batch ADMM has been well-studied \cite{deng2012global,nishihara2015general}, that
on stochastic ADMM is still unclear.
We show that its optimal setting is, interestingly, the same as that in the batch setting.
Moreover,  
SVRG-ADMM 
does not need to store the gradients
or dual variables
throughout the iterations. This makes it particularly appealing when both the number of 
samples and label classes are large. In addition, we also study the convergence properties of the proposed method for nonconvex problems, and obtain a convergence rate of $O(1/T)$ to a stationary point.

\noindent {\bf Notation}:   
For a vector $x$,
$\|x\|$ 
is its $\ell_2$-norm, and 
$\|x\|_Q = \sqrt{x^T Q x}$. 
For a matrix $X$, $\|X\|$ is its spectral norm, 
$\sigma_{\max}(X)$ (resp. $\sigma_{\min}(X)$) is its largest (resp. smallest) eigenvalue,
and $X^{\dagger}$ its pseudoinverse.  
For a function $f$, $f'$ is a subgradient.  When $f$ is differentiable, we use $\nabla f$
as its gradient.

%%%%%%%%%%%%%%%%%%%%%%%%%%%%%%%%%%%%%%%%%%%%%%%%%%%%%%%%%%%%%%%%%%%%

\section{Related Work}
\label{sec:review}

Consider the 
regularized risk minimization problem:
$\min_x \frac 1 n\sum_{i=1}^n f_i(x) +r(x)$, 
where $x$ is the model parameter, 
$n$ is the number of training samples, 
$f_i$ is the loss due to sample $i$,
and
$r$ is a regularizer. For many 
structured sparsity regularizers, $r(x)$ is of the form
$g(Ax)$, where $A$ is a matrix
\cite{kim2009multivariate,jacob2009group}.
By introducing an additional $y$, 
the problem can be rewritten as
\begin{eqnarray} \label{eq:problem_special}
\min_{x,y} \; f(x) + g(y) \;:\; Ax - y = 0, 
\end{eqnarray} 
where 
\begin{equation} \label{eq:f}
f(x)=\frac 1 n\sum_{i=1}^n f_i(x).
\end{equation} 
Problem~(\ref{eq:problem_special}) can be conveniently  solved by 
the 
alternating direction method of multipliers (ADMM)
\cite{boyd2011distributed}.
In general, ADMM considers problems of the form 
\begin{equation} \label{eq:problem}
\min_{x,y} \; f(x) + g(y) \;:\; Ax + By = c, 
\end{equation} 
where $f,g$ are convex functions, and 
$A,B$ (resp. $c$) are constant matrices (resp. vector).
Let $\rho > 0$ be a penalty parameter, and
$u$ be the dual variable.
At iteration $t$,  ADMM performs the updates:
\begin{eqnarray}
y_{t} \!\!\! &=& \!\!\!  \arg \min_{y} g(y) + \frac{\rho}{2}\|Ax_{t-1} + By - c +
u_{t-1}\|^2, \label{eq:classic_y}\\
x_{t} \!\!\! &=& \!\!\! \arg \min_{x} f(x) + \frac{\rho}{2}\|Ax + By_{t} - c + u_{t-1}\|^2, \label{eq:classic_x} \\
u_{t} \!\!\! &=& \!\!\!  u_{t-1} + Ax_{t} + By_{t} - c. \label{eq:classic_u}
\end{eqnarray}

With $f$ in (\ref{eq:f}), solving (\ref{eq:classic_x}) can be computationally expensive when the data set is large.
Recently, 
a number of stochastic and online variants of ADMM have been developed
\cite{Wang2012,ouyang2013stochastic,suzuki2013dual}.
However, 
they converge 
much slower than the batch ADMM, namely,
$O(1/\sqrt{T})$ vs
$O(1/T)$
for convex problems, and $O(\log T/T)$ vs
linear convergence
for strongly convex problems.

For gradient descent,
a similar gap in convergence rates between the stochastic and batch algorithms is well-known
\cite{roux2012stochastic}.
As noted by 
\cite{johnson2013accelerating}, the underlying reason 
is that 
SGD has to 
control the gradient's variance by gradually reducing its
stepsize $\eta$.
Recently, by observing 
that the training set is always finite in practice, a number of 
variance reduction techniques 
have been developed
that 
allow the use of
a constant stepsize, and consequently faster convergence.
In this paper, we focus on  the
SVRG 
\cite{johnson2013accelerating}, 
which is advantageous 
in that no
extra
space for the intermediate gradients or dual variables
is needed.
The algorithm proceeds in stages.
At the beginning of each stage, 
the gradient
$\tz = \frac{1}{n}\sum_{i=1}^n\nabla f_i(\tx)$
is computed 
using a past parameter estimate $\tx$.
For each 
subsequent
iteration $t$ in this stage,
the approximate gradient 
\begin{equation} \label{eq:approx}
\hat{\nabla} f(x_{t-1}) =
%\nabla f_{i_t}(x_{t-1}) - \nabla f_{i_t}(\tx) + \tz 
\frac{1}{b}\sum_{i_t \in \mathcal{I}_t}(\nabla f_{i_t}(x_{t-1}) - \nabla f_{i_t}(\tx)) + \tz 
\end{equation} 
is used,
where $\mathcal{I}_t$ is a mini-batch of size $b$
from $\{1, 2, \dots, n\}$.
Note that $\hat{\nabla} f(x_{t-1})$ is unbiased (i.e., $\CE\hat{\nabla} f(x_{t-1})= \nabla
f(x_{t-1})$), and
its (expected) variance  
goes to zero asymptotically.

Recently, variance reduction 
has also been
incorporated into stochastic ADMM.
For example, SAG-ADMM \cite{zhong2014fast} is based on 
SAG \cite{roux2012stochastic}; and SDCA-ADMM \cite{suzuki2014stochastic} is based on 
SDCA \cite{shalev2013stochastic}.
Both enjoy low iteration complexities and fast convergence.
However, SAG-ADMM requires $O(nd)$ space
for the old gradients and weights,
where $d$ is the dimensionality of $x$.
As for SDCA-ADMM, 
even though its space requirement  is lower, it
is still proportional to $N$,
the number of labels
in a multiclass/multilabel/multitask learning problem.
As $N$ can easily be in the
thousands or even millions
(e.g., Flickr has 
more than 20 millions tags), 
SAG-ADMM and SDCA-ADMM
can still be problematic.

%%%%%%%%%%%%%%%%%%%%%%%%%%%%%%%%%%%%%%%%%%%%%%%%%%%%%%%%%%%%%%%%%%%%%%%%%%%%%%

\section{Integrating SVRG with Stochastic ADMM}
\label{sec:mthd}

In this paper, 
we make the following assumptions on the $f_i$'s in (\ref{eq:f}) and $g$ in (\ref{eq:problem}).

\begin{assumption} \label{assum_smooth}
Each 
%loss function 
$f_i$ 
is convex, continuously
differentiable,
%on $\R^d$, 
and has
$L_i$-Lipschitz-continuous
gradient.
\end{assumption}
Hence, 
for each $i = 1, \dots, n$,
there
exists $L_i > 0$ such that 
\[f_i(x_j) \leq f_i(x_i) + \nabla f_i(x_i)^T(x_j - x_i) + \frac{L_i}{2}\|x_i - x_j\|^2, \forall x_i, x_j\]
%for all $x_i, x_j$.
%\in \R^d$.
%This implies $f$ is also smooth, with $f(x_j) \leq f(x_i) + \nabla f(x_i)^T(x_j - x_i) + \frac{L_f}{2}\|x_i - x_j\|^2$, where $L_f \leq \frac{1}{n}\sum_{i=1}^nL_i \leq \max_i L_i$. 
Moreover, 
Assumption~\ref{assum_smooth}
implies that $f$ is also smooth, with
\[f(x_j) \leq f(x_i) + \nabla f(x_i)^T(x_j - x_i) + \frac{L_f}{2}\|x_i - x_j\|^2,\]
where $L_f \leq \frac{1}{n}\sum_{i=1}^nL_i \leq  \max_i L_i$. Let $L_{\max} = \max_i L_i$. We
thus also have
\[f(x_j) \leq f(x_i) + \nabla f(x_i)^T(x_j - x_i) + \frac{L_{\max}}{2}\|x_i - x_j\|^2.\]

\begin{assumption} \label{assum_g}
$g$ is convex, but can be nonsmooth. 
\end{assumption}

Let $(x_{*}, y_{*})$ be the optimal (primal) solution of (\ref{eq:problem}), and 
$u_{*}$ the corresponding dual solution.
At optimality, we have
\begin{equation} \label{eq:opt1}
\nabla f(x_{*}) + \rho A^Tu_{*} = 0,  \;\;
g'(y_{*}) + \rho B^Tu_{*}  =  0, 
\end{equation} 
\begin{equation} \label{eq:opt3}
Ax_{*} + By_{*}  = c.
\end{equation}

%%%%%%%%%%%%%%%%%%%%%%%%%%%%%%%%%%%%%%%%%%

\subsection{Strongly Convex Problems}

In this section, we consider the case where $f$ is strongly convex.  A popular example in machine learning is the square loss. 

\begin{assumption} \label{assum_scvx} 
$f$ is strongly convex, i.e., there exists $\lambda_f >
0$ such that 
$f(x_i) \geq f(x_j) + \nabla f(x_j)^T(x_i - x_j) + \frac{\lambda_f}{2}\|x_i - x_j\|^2$
for all $x_i,x_j$.
\end{assumption}

Moreover, we assume that matrix $A$ has full row rank.
This assumption has been commonly used
in 
the convergence analysis of ADMM algorithms
\cite{deng2012global,ghadimi-14,giselsson2014diagonal,nishihara2015general}.

\begin{assumption} \label{assum_A}
Matrix $A$ has full row rank.
\end{assumption}

The proposed procedure
is shown in Algorithm~\ref{alg:admm_svrg}.
Similar to SVRG, it is divided
into stages, each with $m$
iterations. 
The updates for $y_t$ and $u_t$ are the same as batch ADMM
((\ref{eq:classic_y})
and (\ref{eq:classic_u})).
The key change is on the more expensive $x_t$ update. We  first
replace (\ref{eq:classic_x})
by its first-order approximation $f(x_{t-1})+\nabla f(x_{t-1})^Tx$.
As in SVRG, the full gradient $\nabla f(x_{t-1})$ is approximated
by $\hat{\nabla} f(x_{t-1})$ in (\ref{eq:approx}).
Recall that $\hat{\nabla} f(x_{t-1})$ is unbiased 
and its (expected) variance  goes to zero. In other words,
$\hat{\nabla} f(x_{t-1}) \rightarrow \nabla f(x_{*})$ when $x_{t-1}$ and $\tx$ approach the
optimal $x_*$,
which allows the use of a constant stepsize.
In contrast, traditional stochastic approximations such as OPG-ADMM
\cite{suzuki2013dual} use $\frac{1}{b}\sum_{i_t \in \mathcal{I}_t}\nabla f_{i_t}(x_{t-1})$
to approximate the full gradient, and 
%it will not go to $\nabla f(x_{*})$ when $x_{t-1} \rightarrow x_*$, thus, 
a decreasing step size is needed to ensure convergence.
%which can dramatically slow down convergence. 

Unlike SVRG, the optimization subproblem in
Step~9  
%of Algorithm~\ref{alg:admm_svrg}
has the additional terms $\frac{\rho}{2}\|Ax + By_t - c + u_{t-1}\|^2$ (from subproblem
(\ref{eq:classic_x})) and
$\frac{1}{2\eta}\|x - x_{t-1}\|_G^2$ (to ensure that the next iterate is close to the
current iterate
$x_{t-1}$).
%which is scaled by $G \succeq I$. 
A common setting for $G$ is simply $G = I$ \cite{ouyang2013stochastic}.
Step~9  
then reduces to
\begin{eqnarray}
x_t \!\!\!\!\! & = & \!\!\!\!\! 
\left(\frac{1}{\eta}I + \rho A^TA\right)^{-1} 
\nonumber \\
& & \!\!\!\!\! \left(\frac{x_{t-1}}{\eta} \!\!-\!\! \hat{\nabla} f(x_{t-1})
+ \rho A^T(By_{t} - c + u_{t-1})
\right). \label{eq:svrg_ex} 
\end{eqnarray}
Note that 
$(\frac{1}{\eta}I
+ \rho A^TA)^{-1}$ above
can be pre-computed. 
On the other hand, 
while some stochastic ADMM algorithms \cite{ouyang2013stochastic,zhong2014fast} also need to compute 
a similar matrix inverse, 
their $\eta$'s change with iterations and so cannot be pre-computed. 

When $A^TA$ is large,
storage of this matrix may still be problematic.
To alleviate this, a common approach is {\em linearization\/} (also called the inexact Uzawa
method) \cite{zhang2011unified}. It sets
$G = \gamma I - \eta\rho A^TA$ 
with
\begin{equation} \label{eq:smallest}
\gamma\geq \gamma_{\min} \equiv \eta\rho\|A^TA\| + 1
\end{equation} 
to ensure that $G \succeq I$.
The $x_t$ update 
in (\ref{eq:svrg_ex})
then simplifies to
\begin{eqnarray}
x_t & = & x_{t-1} - \frac{\eta}{\gamma}\left(
\hat{\nabla} f(x_{t-1}) \right. \nonumber\\ 
&&\left. 
+ \rho A^T(Ax_{t-1} + By_{t} - c + u_{t-1})\right). \label{eq:svrg_iu}
\end{eqnarray}

\begin{algorithm}[th]
   \caption{
	SVRG-ADMM for strongly convex problems.}
   \label{alg:admm_svrg}
\begin{algorithmic}[1]
   \STATE {\bfseries Input:} $m, \eta, \rho > 0.$	
   \STATE initialize 
	$\tx_0, \ty_0$  and $\tu_0 = -\frac{1}{\rho}(A^T)^{\dagger}\nabla f(\tx_{0})$;
   \FOR{$s=1, 2, \dots$}
   \STATE $\tx = \tx_{s - 1}$;
   \STATE{$x_{0} = \tx_{s - 1}$};
   {$y_{0} = \ty_{s - 1}$};
   {$u_{0} = \tu_{s - 1}$};
   \STATE{$\tz = \frac{1}{n}\sum_{i=1}^n\nabla f_i(\tx)$};
   \FOR{$t=1, 2, \dots, m$}
   \STATE{$y_{t} \leftarrow  \arg \min_{y} g(y) + \frac{\rho}{2}\|Ax_{t - 1} + By - c +
	u_{t-1}\|^2$};
   \STATE{$x_{t} \leftarrow  \arg \min_{x} 
\hat{\nabla} f(x_{t-1})^Tx + \frac{\rho}{2} \|Ax + By_{t} - c + u_{t-1}\|^2 + \frac{\|x - x_{t-1}\|^2_{G}}{2\eta}
	$};
   \STATE{$u_{t}  \leftarrow  u_{t-1} + Ax_{t} + By_{t} - c  $};
   \ENDFOR
   \STATE{$\tx_{s} = \frac{1}{m}\sum^{m}_{t=1}x_t$};
   {$\ty_{s} = \frac{1}{m}\sum^{m}_{t=1}y_t$};
   {$\tu_{s} = -\frac{1}{\rho}(A^T)^{\dagger}\nabla f(\tx_{s})$}; 
   \ENDFOR
   \STATE {\bfseries Output:} 
   $\tx_{s}, \ty_{s}$;
\end{algorithmic}
\end{algorithm}

Note that steps~2 and 12 in Algorithm~\ref{alg:admm_svrg}
involve the pseudo-inverse $A^{\dagger}$.
As $A$ is often sparse,
this can be efficiently computed 
by the Lanczos algorithm \cite{golub2012matrix}.

In general,
as in other stochastic algorithms, 
the stochastic gradient is computed based on a mini-batch  of size $b$.
The following Proposition shows that the variance can be progressively reduced.
Note that 
this and other
results in this section also hold for the batch mode, in which
the whole data set is used in each iteration
(i.e., $b=n$).
\begin{prop} \label{lemma:variance}
The variance of 
$\hat{\nabla} f(x_{t-1})$
is bounded by
$\CE\|
\hat{\nabla} f(x_{t-1})
- \nabla f(x_{t-1})\|^2
\leq 4L_{\max}\beta(b)\left(J(x _{t- 1}) - J(x _{*}) + J(\tx) - J(x _{*})\right)$,
where 
$L_{\max} \equiv \max_i L_i$,
$\beta(b) = \frac{n-b}{b(n-1)}$, 
$J(x) = f(x) + \rho u_{*}^TAx$, and
$J(x _{t- 1}) - J(x _{*}) + J(\tx) - J(x _{*})\geq 0$.
\end{prop}

Using (\ref{eq:opt1}) and (\ref{eq:opt3}),
$J(x) - J(x_{*}) = f(x) - f(x_{*}) - \nabla f(x_{*})^T(x - x_{*}) = 0$ when $x \rightarrow x_{*}$, and thus the variance goes to zero.
Moreover, 
as expected, the variance reduces when $b$ increases, and goes to zero when $b = n$. However, a
large $b$ leads to a high per-iteration cost. Thus, there is a tradeoff between ``high
variance with cheap iterations" and ``low variance with expensive iterations".

%%%%%%%%%%%%%%%%%%%%%%%%%%%%%%%%%%%%%%%%%%

\subsubsection{Convergence Analysis}

%%%%%%%%%%%%%%%%%%%%%

In this section, we study
the convergence
w.r.t. $R(x, y) \equiv f(x) - f(x_{*}) - \nabla f(x_{*})^T(x - x_{*}) 
+ g(y) - g(y_{*}) - g'(y_{*})^T(y - y_{*})$.
First, note that 
$R(x, y)$
is always non-negative.

\begin{prop}
$R(x, y) \geq 0$ for any $x$ and $y$. 
\end{prop}
Using the optimality conditions in (\ref{eq:opt1}) and (\ref{eq:opt3}),
$R(x, y)$
can be rewritten as
$f(x) + g(y) + \rho u_{*}^T(Ax + By -c) 
- (f(x_{*}) + g(y_{*}) + \rho u_{*}^T(Ax_{*} + By_{*} -c))$,
which is the difference of the Lagrangians in (\ref{eq:problem}) evaluated at $(x, y, u_{*})$ and $(x_{*}, y_{*}, u_{*})$.
Moreover,
$R(x, y) \geq 0$
is the same as the variational inequality
used in \cite{he20121}.

The following shows that Algorithm~\ref{alg:admm_svrg} converges linearly.

\begin{theorem} \label{convergence_theorem}
Let
\begin{eqnarray}
\kappa  \!&\!  =  \! & \! \frac{\|G + \eta\rho A^TA\|}{\lambda_f\eta(1 - 4L_{\max}\eta\beta(b))m} +
\frac{4L_{\max}\eta\beta(b)(m + 1)}{(1 - 4L_{\max}\eta\beta(b))m}  \nonumber\\ 
&& + \frac{L_f}{\rho(1 - 4L_{\max}\eta\beta(b))\sigma_{\min}(AA^T)m}. \label{eq:kappa}
\end{eqnarray}
Choose $0 < \eta < \min\left\{\frac{1}{L_f}, \frac{1}{4L_{\max}\beta(b) }\right\}$,  
and  the number of iterations $m$ is sufficiently large such that 
$\kappa  <  1$.
Then, 
$\CE R(\tx_s, \ty_s) \leq  \kappa^s  R(\tx_{0}, \ty_{0})$.
\end{theorem}
 
Theorem~\ref{convergence_theorem} 
is similar to the SVRG results in \cite{johnson2013accelerating,xiao2014proximal}.
However, 
it is not a trivial extension 
because of the presence of the equality constraint and Lagrangian multipliers in the ADMM formulation.
Moreover, for the existing stochastic ADMM algorithms, linear convergence is only proved in SDCA-ADMM
for a special case
($B=-I$
and $c = 0$
in (\ref{eq:problem})).
Here, we have linear
convergence for a general $B$ and any $G \succeq I$ (in step~9).
%and is more
%general than the results in 
%\cite{ouyang2013stochastic,suzuki2013dual,suzuki2014stochastic,zhong2014fast}.

\begin{corollary} \label{cor:prob_bnd}
For a fixed $\k$ and $\epsilon>0$,
the number of stages $s$ 
required to ensure $\CE R(\tx_s, \ty_s) \leq \epsilon$ 
is
$s \geq \log \left(\frac{R(\tx_0, \ty_0)}{\epsilon}\right)/\log \left(\frac{1}{\kappa}
\right)$. Moreover, 
for any $\delta \in (0, 1)$,
we have the high-probability bound: 
$\mathrm{Prob}(R(\tx_s, \ty_s) \leq \epsilon) \geq 1 - \delta$ if 
$s \geq \log \left(\frac{R(\tx_0, \ty_0)}{\epsilon\delta}\right)/\log \left(\frac{1}{\kappa}
\right)$.
\end{corollary}

\subsubsection{Optimal ADMM Parameter
$\rho$}

With linearization,
the first term in 
(\ref{eq:kappa}) becomes $\|\gamma I\|/(\lambda_f\eta(1 -
4L_{\max}\eta\beta(b))m)$. Obviously, it is desirable to have a small convergence factor $\k$, and so we will
always use $\gamma = \gamma_{\min}$
in (\ref{eq:smallest}). 
The following Proposition obtains the optimal $\rho_*$,
which yields the smallest $\kappa$ value
and thus fastest convergence. 
Interestingly, 
this $\rho_*$ is the same
as that of its batch counterpart (Theorem $7$ in \cite{nishihara2015general}). 
In other words, the optimal $\rho_*$
is not affected by the stochastic approximation. 

\begin{prop} \label{prop:kappa}
%For both $x_t$ updates in (\ref{eq:svrg_ex}) and (\ref{eq:svrg_iu}), 
%With $\gamma=\gamma_{\min}$ in (\ref{eq:smallest}),
Choosing
\begin{equation} \label{eq:rho}
\rho = \rho_{*} \equiv \sqrt{\frac{L_f
\lambda_f}{\sigma_{\max}(AA^T)\sigma_{\min}(AA^T)}}
\end{equation} 
yields the smallest $\kappa$:
%in (\ref{eq:kappa}):
\begin{eqnarray} 
\kappa_{\min} \!\!\!\! & =  & \!\!\!\! \frac{1}{\lambda_f\eta(1 -
4L_{\max}\eta\beta(b))m} + \frac{4L_{\max}\eta\beta(b)(m + 1)}{(1 -
4L_{\max}\eta\beta(b))m}  \nonumber \\
& & \!\!\!\! + \frac{2h_A\sqrt{h_f}}{(1 - 4L_{\max}\eta\beta(b))m}, \label{eq:skappa}
\end{eqnarray}
where $h_f = \frac{L_f}{\lambda_f}$ 
is the 
%smallest 
condition number of $f$, 
and $h_A 
= \sqrt{\frac{\sigma_{\max}(AA^T)}{\sigma_{\min}(AA^T)}}$
%= \frac{\sigma_{\max}(A)}{\sigma_{\min}(A)}$
is the condition number of $A$. 
\end{prop}

Assume that we have a target value for $\kappa$, say,
$\tkappa$ (where
$\kappa_{\min}
\leq \tkappa <1$). 
Let 
$\eta_{*}$ 
be the 
$\eta$ value
that minimizes
%\footnote{\#*** however, just min the number of inner iterations may not be enough, as it may lead to more stages. one would like to min the total computational cost. No, it won' lead to more stages. Here we have a target $\tkappa$, so it means each stage reduce the same amount of objective. Then, minimizing the cost in each stage will lead to less overall cost.}
the number of inner iterations ($m_{*}$) in Algorithm~\ref{alg:admm_svrg} while still achieving the
target $\tkappa$.

\begin{prop}  \label{prop_opt_m}
Fix $\rho=\rho_*$, and
define 
\begin{eqnarray} \label{eq:opt_eta}
\tilde{\eta} &  =  &  \sqrt{\left(\frac{1+\tkappa}{\tkappa\lambda_f + 2(1+\tkappa)\sqrt{L_f\lambda_f}h_A}\right)^2 + \delta} \nonumber \\
&& - \frac{1 +\tkappa}{\tkappa\lambda_f + 2(1+\tkappa)\sqrt{L_f\lambda_f}h_A},
\end{eqnarray}
where $\delta = \frac{1}{4L_{\max}\lambda_f\beta(b)(1 +
2(1+1/\tkappa)h_A\sqrt{h_f})}$. 
\begin{enumerate}
\item If $b \leq b_{*} \equiv \frac{n}{M(n-1) +1}$
where $M= \frac{\tkappa h_f L_f / L_{\max}}{8((1 + \tkappa)(h_f + h_A\sqrt{h_f}) +
\tkappa/2)}$, then 
\begin{eqnarray}
\eta_{*} \!\!\!\! & = & \!\!\!\! \tilde{\eta} \leq \frac{1}{L_f}, \nonumber \\
m_{*} \!\!\!\! &  =& \!\!\!\! \frac{8\beta(b)h_Q}{\tkappa^2} 
\left(\!\! \sqrt{(1 + \tkappa)^2 +
\frac{\tkappa^2}{16\beta(b)^2L_{\max}^2\delta}}\!\! + \!\! 1 \!\! + \!\! \tkappa \right)
\nonumber\\
&& \!\!\!\! + \frac{2h_A\sqrt{h_f}}{\tkappa}, \label{eq:m_opt_1}
\end{eqnarray}
where
$h_Q = \frac{L_{\max}}{\lambda_f}$.
%\geq h_f$.
%is also the condition number of the loss function $f$. 
\item Otherwise, 
\begin{eqnarray}
\eta_{*} & = & \frac{1}{L_f}, \nonumber\\
m_{*} &  =  &  \frac{h_f + 4\beta(b)L_{\max}/L_f + 2h_A\sqrt{h_f}}{\tkappa - (1 +
\tkappa)4\beta(b)L_{\max}/L_f}.  \label{eq:m_opt_2}
\end{eqnarray}
\end{enumerate}
\end{prop} 

\begin{remark}
As expected, if the target $\tkappa$ is very small, $m_*$ can be large. 
It is also easy to see from (\ref{eq:m_opt_1}) and (\ref{eq:m_opt_2}) that $m_{*}$ 
%monotonically 
decreases w.r.t. $b$,
and increases with $h_f$ and $h_A$.
%Moreover, $m_* = O(\max(h_A\sqrt{h_f}, h_Q))$, and so convergence becomes slower with larger condition numbers of $f$ and $A$. 
\end{remark}

%%%%%%%%%%%%%%%%%%%%%%%%%%%%%%%%%%%%%%%%%%%%%%%%%%%%%%%%

\subsection{General Convex Problems}

In this section, we consider
(general) convex problems, and 
only Assumptions~\ref{assum_smooth}, \ref{assum_g} are needed.
The procedure (Algorithm~\ref{alg:admm_svrg_general})
differs slightly from Algorithm~\ref{alg:admm_svrg}
in the initialization of each stage 
(steps 2, 5, 12)
and the final output (step 14). 

As expected, with a weaker form of convexity, the convergence rate 
of Algorithm~\ref{alg:admm_svrg_general}
is no longer linear.
Following \cite{ouyang2013stochastic,suzuki2013dual,zhong2014fast},
we consider the convergence of
$R(\bar{x}, \bar{y}) + \zeta\|A\bar{x} + B\bar{y} - c\|$, where
$\zeta > 0$ and
$\|A\bar{x} + B\bar{y} - c\|$ measures the feasibility of the ADMM solution.
The following Theorem shows that 
Algorithm~\ref{alg:admm_svrg_general}
has 
$O(1/s)$ convergence.
Since both $R(\bar{x}, \bar{y})$ and $\|A\bar{x} + B\bar{y} - c\|$ are always nonnegative, 
obviously each term individually also has $O(1/s)$ convergence.

\begin{algorithm}[ht]
   \caption{
	SVRG-ADMM for general convex 
	problems.}
   \label{alg:admm_svrg_general}
\begin{algorithmic}[1]
   \STATE {\bfseries Input:} $m, \eta, \rho > 0.$
   \STATE initialize 
   $\tx_0 = \hx_{0}, \hy_0$  and $\hu_0$;
   \FOR{$s=1, 2, \dots$}
   \STATE $\tx = \tx_{s - 1}$;
   \STATE{$x_{0} = \hx_{s - 1}$};
   {$y_{0} = \hy_{s - 1}$};
   {$u_{0} = \hu_{s - 1}$};
   \STATE{$\tz = \frac{1}{n}\sum_{i=1}^n\nabla f_i(\tx)$};
   \FOR{$t=1, 2, \dots, m$}
   \STATE{$y_{t} \leftarrow  \arg \min_{y} g(y) + \frac{\rho}{2}\|Ax_{t - 1} + By - c +
	u_{t-1}\|^2$};
   \STATE{$x_{t} \leftarrow  \arg \min_{x} \hat{\nabla} f(x_{t-1})^Tx + \frac{\rho}{2} \|Ax + By_{t} - c + u_{t-1}\|^2 + \frac{\|x - x_{t-1}\|^2_{G}}{2\eta} $};
   \STATE{$u_{t}  \leftarrow  u_{t-1} + Ax_{t} + By_{t} - c  $};
   \ENDFOR
   \STATE{$\tx_{s} = \frac{1}{m}\sum^{m}_{t=1}x_t$};
   {$\ty_{s} = \frac{1}{m}\sum^{m}_{t=1}y_t$};
   {$\hx_{s} = x_m$};
   {$\hy_{s} = y_m$};
   {$\hu_{s} = u_m$}; 
   \ENDFOR
   \STATE {\bfseries Output:} 
    $\bar{x} = \frac{1}{s}\sum_{i=1}^s\tx_i, \bar{y} =\frac{1}{s}\sum_{i=1}^s\ty_{s}$.
\end{algorithmic}
\end{algorithm}
%%%%%%%%%%%%%%%%%%%%%%%%%%%%%%%%%%%%%%%%%%%%%%%%%%%%%%%%

\begin{theorem} \label{convergence_theorem_nonstrong}
Choose $0 < \eta < \min\left\{\frac{1}{L_f}, \frac{1}{8L_{\max}\beta(b) }\right\}$. Then, 
\begin{eqnarray}
\lefteqn{\CE (R(\bar{x}, \bar{y}) + \zeta\|A\bar{x} + B\bar{y} - c\|)} \nonumber\\
& \!\!\!\!\!\!\!\!  \leq \!\!\! & \!\!\!\!  \frac{4L_{\max}\eta\beta(b)(m\!\!+\!\!1)}{(1 \!\!-\!\! 8L_{\max}\eta\beta(b))ms}\left(f(\hx _{0}) \!\!- \!\! f(x_{*}) \!\! - \!\! \nabla f(x _{*})^T(\hx _{0} \!\!-\!\! x _{*})\right) \nonumber\\
&& \!\!\!\!  + 
\frac{\frac{1}{2\eta}\|\hx _{0} - x _{*}\|^2_{G + \eta \rho A^TA} + \rho\left(\|\hu_{0} - u_{*}\|^2 +
\frac{\zeta^2}{\rho^2} \right)}{(1 - 8L_{\max}\eta\beta(b))ms}. \label{eq:sublinear}
\end{eqnarray}
\end{theorem}

The following Corollary obtains a sublinear convergence rate for the batch case
($b = n$).
This is similar to that of
Remark~1 in \cite{ouyang2013stochastic}. However, here we allow a general $G$ while they
require $G = I$. 

\begin{corollary} \label{cor:batch}
In batch learning,  
\begin{eqnarray}\label{eq:batch}
\lefteqn{R(\bar{x}, \bar{y}) +
\zeta\|A\bar{x} + b\bar{y} - c\|} \nonumber\\
& \leq & \frac{1}{2\eta ms}\|\tx _{0} - x _{*}\|^2_{G + \eta \rho A^TA}
+
\frac{\rho}{ms} \left(\|\tu_{0} - u_{*}\|^2 +
\frac{\zeta^2}{\rho^2} \right).
\end{eqnarray}
\end{corollary}

\begin{remark}
When $b = n$, the whole data set is used in each iteration, and $\frac{1}{b}\sum_{i_t \in
\mathcal{I}_t}(\nabla f_{i_t}(x_{t-1}) - \nabla f_{i_t}(\tx)) + \tz$ in the $x_t$ update 
reduces to $\frac{1}{n}\sum_{i=1}^n\nabla f_{i}(x_{t-1})$. Each iteration is then simply standard
batch ADMM (with linearization), and the whole procedure is the same as running batch ADMM
for a total of $ms$ iterations. Not surprisingly,
the RHS in (\ref{eq:batch}) can still go to zero by just
setting $m = 1$ (with increasing $s$) or $s = 1$ (with increasing $m$).
In contrast,  when
$b\neq n$,
%this is not possible 
%one cannot have convergence by only increasing $m$ and setting $s = 1$.
setting $s = 1$ 
in (\ref{eq:sublinear})
cannot guarantee convergence. Intuitively,
the past full gradient used in that single stage is only an approximation of the  batch
gradient, and the variance of the stochastic gradient cannot be reduced to zero.  On the other hand, if each stage has only one iteration ($m = 1$), we have
% in (\ref{eq:sublinear}), 
 $x_0 = \tx$, and
 $\frac{1}{b}\sum_{i_t \in
 \mathcal{I}_1}(\nabla f_{i_1}(x_{0}) - \nabla f_{i_t}(\tx)) + \tz$ in the $x_1$ update
 reduces to $\tz$.
Thus, it is the same as batch ADMM with a total of $s$ iterations.
\end{remark}

%%%%%%%%%%%%%%%%%%%%%%%%%%%%%%%%%%%%%%%%%%%%%%%%%%%%%%%%%%%%%%%%%%%%%%%%%%%%%%

\subsection{Nonconvex Problems}

In this section, we consider nonconvex problems. The algorithm is shown in
Algorithm~\ref{alg:admm_svrg_nc}. Let $g_* = \inf_y g(y) > -\infty$, and $f_* = \inf_x f(x) > -\infty$.
Moreover, we also use Assumptions~\ref{assum_g}, \ref{assum_A} and the following.

\begin{algorithm}[tb]
   \caption{
	SVRG-ADMM for nonconvex problems.}
   \label{alg:admm_svrg_nc}
\begin{algorithmic}[1]
   \STATE {\bfseries Input:} $m, \eta, \rho > 0.$	
   \STATE initialize 
	$\tx_0, \ty_0$  and $\tu_0$;
   \FOR{$s=1, 2, \dots, S$}
   \STATE $\tx = \tx_{s - 1}$;
   \STATE{$x_{0} = \tx_{s - 1}$};
   {$y_{0} = \ty_{s - 1}$};
   {$u_{0} = \tu_{s - 1}$};
   \STATE{$\tz = \frac{1}{n}\sum_{i=1}^n\nabla f_i(\tx)$};
   \FOR{$t=1, 2, \dots, m$}
      \STATE{$y_{t} \leftarrow  \arg \min_{y} g(y) + \frac{\rho}{2}\|Ax_{t-1} + By - c +
	u_{t-1}\|^2$};
   \STATE{$x_{t} \leftarrow  \arg \min_{x} 
\hat{\nabla} f(x_{t-1})^Tx + \frac{\rho}{2} \|Ax + By_{t} - c + u_{t-1}\|^2 + \frac{\|x - x_{t-1}\|^2_{G}}{2\eta}
	$};
   \STATE{$u_{t}  \leftarrow  u_{t-1} + Ax_{t} + By_{t} - c  $};
   \ENDFOR
   \STATE{$\tx_{s} = x_m$};
   {$\ty_{s} = y_m$};
   {$\tu_{s} = u_m$}; 
   \ENDFOR
   \STATE {\bfseries Output:} 
   Iterate $(x_o, y_o)$ chosen uniformly at random from ($\{\{x^s_t\}_{t=1}^{m}\}_{s=1}^{S}$, $\{\{y^s_t\}_{t=1}^{m}\}_{s=1}^{S}$);
\end{algorithmic}
\end{algorithm}

\begin{assumption} \label{assum_smoothnc}
Each $f_i$ is
continuously differentiable has $L_i$-Lipschitz-continuous gradient,
and possibly nonconvex.
\end{assumption}
%Below we provide a nonconvex smooth function $f_i(x)$. This nonconvex loss function will be used in our experiment.
As an example, the sigmoid loss function,
$f_i(x) = (1 + \exp(o_iz_i^Tx))^{-1} \in [0, 1]$, where $o_i \in \{-1, 1\}$ is the label and $z_i$ is
the feature vector,
satisfies Assumption~\ref{assum_smoothnc}. In this case,
we have $\|\nabla f_i(x)\| = \left\|\frac{\exp(o_iz_i^Tx)}{(1 + \exp(o_iz_i^Tx))^2}o_iz_i\right\| \leq \frac{1}{4}\|z_i\|$.

Define the augmented Lagrangian function
\[L(x, y, u) = f(x) + g(y) + \rho u^T(Ax + By -c) + \frac{\rho}{2}\|Ax + By - c\|^2.\]
Moreover, define the proximal gradient of the augmented Lagrangian
function as
\begin{eqnarray*}
\tilde{\nabla} L(x, y, u) = \left[\begin{matrix} \nabla_{x}L(x, y, u) \\ 
y - \text{prox}_{g}\left(y - \nabla_{y}\left(L(x, y, u) - g(y)\right)\right) \\
Ax + By - c \end{matrix}\right],
\end{eqnarray*}
where $\text{prox}_g(q) = \min_y g(y) + \frac{1}{2}\|y - q\|^2$. The quantity $\|\tilde{\nabla} L(x, y, u)\|^2$ will be used to measure progress of the algorithm. This is also used 
in \cite{hong2016convergence}
for analyzing the iteration complexity of the vanilla nonconvex ADMM.

\begin{theorem} \label{convergence_theorem_nc}
Choose $0 < \eta < \frac{1}{2L_f}$ small enough and $\rho \geq \frac{4L_f}{\sigma_{min}(AA^T)}$ large
enough so that the following condition holds:
\begin{eqnarray} \label{eq:nc_condition}
& 8L_{max}^2m^2\beta(b)\eta^2 + L_{max}\eta + \frac{36\|G\|}{\eta\rho\sigma_{min}(AA^T)} + \frac{36L_{max}\sqrt{\|G\|}}{\rho\sigma_{min}(AA^T)} \nonumber\\
& + \left(\frac{288L_{max}^2m^2}{\sigma_{min}(AA^T)} + \frac{216L_{max}^2(m + 1)}{\sigma_{min}(AA^T)} + \frac{18L_{max}^2}{\sigma_{min}(AA^T)}\right)\frac{\eta}{\rho}  \leq 1.
\end{eqnarray}
Let $T=mS$. Then, 
\begin{eqnarray*}
\lefteqn{\CE\|\tilde{\nabla} L(x_o, y_o, u_o)\|^2} \nonumber \\
&\leq&  \!\!\!\! \frac{C}{T}\left[L(\tx_{0}, \ty_{0}, \tu_{0}) + \frac{3}{\rho\sigma_{min}(AA^T)}\|\nabla f(\tx_0) + \rho A^T\tu_0\|^2 - \tilde{L}\right],
\end{eqnarray*}
 where $L(\tx_{0}, \ty_{0}, \tu_{0}) + \frac{3}{\rho\sigma_{min}(AA^T)}\|\nabla f(\tx_0) + \rho A^T\tu_0\|^2 \geq \tilde{L} = f_* + g_*$, $C = C_1/C_2$, $C_1 = \max(3(L_f + \rho\|A^TA\|)^2 + 2\rho^2\|B^TA\|^2, \frac{3}{\eta^2}\|G - \eta\rho A^TA\|^2, 3\rho^2\|A\|^2 + 2\rho^2\|B\|^2 + 1)$ and $C_2 = \min\left(\frac{1}{2\eta} - L_f, \frac{1}{4\eta}, \frac{\rho}{2}\right)$. 
%Furthermore, the sequence $\CE[\|x_o - x_{o-1}\|^2_G + \|y_o - y_{o-1}\|^2_{B^TB} + \|u_o - u_{o-1}\|^2] \leq O(1/T)$.
\end{theorem}

When Assumption~\ref{assum_g} does not hold, $g$ can be nonsmooth and nonconvex. In this case, $\partial
g$ denotes the general subgradients of $g$ (Definition 8.3 in \cite{rockafellar2009variational}). We use general subdifferential
\begin{eqnarray*}
\partial L(x, y, u) = \left[\begin{matrix} \nabla_{x}L(x, y, u) \\ 
\partial L_y(x, y, u) \\
Ax + By - c \end{matrix}\right].
\end{eqnarray*}

\begin{theorem} \label{convergence_theorem_ncg}
If $g$ is possibly nonconvex, choose $0 < \eta < \frac{1}{2L_f}$ small enough and $\rho \geq \frac{4L_f}{\sigma_{min}(AA^T)}$ large enough so that (\ref{eq:nc_condition}) holds. 
Let $T=mS$. Then, 
\begin{eqnarray*}
\lefteqn{\CE [\text{dist}(0, \partial L(x, y, u))]^2} \nonumber \\
&\leq&  \!\!\! \frac{C}{T}\left[L(\tx_{0}, \ty_{0}, \tu_{0}) + \frac{3}{\rho\sigma_{min}(AA^T)}\|\nabla f(\tx_0) + \rho A^T\tu_0\|^2 - \tilde{L}\right],
\end{eqnarray*}
where $\text{dist}(0, \partial L(x, y, u))$ is the distance between $0$ and the general subdifferential $\partial L(x, y, u)$, i.e.,
\[\text{dist}(0, \partial L(x, y, u)) = \min_{L'(x, y, u) \in \partial L(x, y, u)}\|0 - L'(x, y, u)\|.\]
%
%Furthermore, the sequence $\CE[\|x_o - x_{o-1}\|^2_G + \|y_o - y_{o-1}\|^2_{B^TB} + \|u_o - u_{o-1}\|^2] \leq O(1/T)$.
\end{theorem}

%%%%%%%%%%%%%%%%%%%%%%%%%%%%%%%%%%%%%%%%%%

\subsection{Comparison with SCAS-ADMM}
\label{sec:discussion}

\begin{table*}[ht]
\caption{Convergence rates and space requirements of 
various stochastic ADMM algorithms, including
stochastic ADMM (STOC-ADMM) 
\protect\cite{ouyang2013stochastic}, 
online proximal gradient descent ADMM (OPG-ADMM) 
\protect\cite{suzuki2013dual}, 
regularized dual averaging ADMM (RDA-ADMM) 
\protect\cite{suzuki2013dual}, 
stochastic averaged gradient ADMM (SAG-ADMM) 
\protect\cite{zhong2014fast}, 
stochastic dual coordinate ascent ADMM  (SDCA-ADMM) 
\protect\cite{suzuki2014stochastic}, scalable stochastic ADMM (SCAS-ADMM)
\protect\cite{zhao2015scalable}, and the proposed SVRG-ADMM.
Here, $d,\tilde{d}$ are dimensionalities of $x$ and $y$ in (\ref{eq:problem}).}
\label{com_alg}
\begin{center}
\begin{tabular}{ccccc} \hline
& general convex & strongly convex & nonconvex & space requirement \\ \hline
STOC-ADMM & $O(1/\sqrt{T})$  &  $O(\log T/T)$  & unknown & $O(d \tilde{d} + d^2)$  \\
OPG-ADMM   & $O(1/\sqrt{T})$  &  $O(\log T/T)$ & unknown &  $O(d \tilde{d})$ \\
RDA-ADMM   & $O(1/\sqrt{T})$  &  $O(\log T/T)$ & unknown  &  $O(d \tilde{d})$ \\ \hline
SAG-ADMM   & $O(1/T)$  &  unknown & unknown &  $O(d \tilde{d} + nd)$  \\
SDCA-ADMM &  unknown  & linear rate & unknown &   $O(d \tilde{d}+ n)$ \\
SCAS-ADMM & $O(1/T)$ & $O(1/T)$ & unknown &   $O(d \tilde{d})$ \\
SVRG-ADMM  &   $O(1/T)$ & linear rate & $O(1/T)$ & $O(d \tilde{d})$
\\\hline
\end{tabular}
\end{center}
\end{table*}

The recently proposed SCAS-ADMM 
\cite{zhao2015scalable}
is a more rudimentary 
integration of SVRG and ADMM.
%Because of the lack of space, we will focus on 
%their algorithm~1 for
%general convex problems.
The main difference with our method is that 
SCAS-ADMM moves the updates of $y$ and $u$ outside the inner {\bf for} loop.
As such, the inner {\bf for} loop focuses only on 
updating
$x$,
and is the same as using a one-stage SVRG to solve for an inexact $x$ solution in
(\ref{eq:classic_x}).
Variables $y$ and $u$ are not updated until 
the $x$ subproblem has been approximately solved
(after running $m$ updates of $x$).

In contrast,
we replace the $x$ subproblem in
(\ref{eq:classic_x})
with its first-order stochastic approximation,
and then
update 
$y$ and $u$ 
in every iteration as $x$.
This difference is analogous to that between the Jacobi iteration and
Gauss-Seidel iteration.
The use of first-order stochastic approximation 
has also shown clear speed advantage in other stochastic
ADMM algorithms \cite{ouyang2013stochastic,suzuki2013dual,zhong2014fast,suzuki2014stochastic},
and is especially desirable on big data sets.

As a result, the convergence rates of SCAS-ADMM are inferior to those of SVRG-ADMM.
On strongly convex problems,
SVRG-ADMM attains a linear convergence rate,
while
SCAS-ADMM 
only has 
$O(1/s)$ 
convergence. On general convex problems, both SVRG-ADMM and SCAS-ADMM have a
convergence rate of $O(1/s)$.  However, 
SCAS-ADMM requires the stepsize to be gradually reduced as $O(1/s^\delta)$, where $\delta>1$.
This defeats the original purpose of using SVRG-based algorithms
(e.g., SVRG-ADMM), which aims at using a
constant learning rate for faster  convergence \cite{johnson2013accelerating}. Moreover,
(\ref{eq:sublinear}) shows that our rate consists of three components, which converge as
$O(1/s)$, $O(1/(ms))$ and $O(1/(ms))$, respectively. On the other hand, while
the sublinear convergence bound in SCAS-ADMM also has three similar components, they all converge
as $O(1/s)$.  
To make the cost of full gradient computation less pronounced,
a natural choice for $m$ is 
$m = O(n)$
\cite{johnson2013accelerating}.
Hence, SCAS-ADMM
can be much slower than SVRG-ADMM when $n$ is large.  
 
%%%%%%%%%%%%%%%%%%%%%%%%%%%%%%%%%%%%%%%%%%

\subsection{
Space Requirement }
\label{sec:space}

The space requirements 
of Algorithms~\ref{alg:admm_svrg} and \ref{alg:admm_svrg_general}
mainly come from step~12.  For simplicity, we consider $B = -I$ and $c = 0$, which are assumed in 
\cite{suzuki2013dual,suzuki2014stochastic}.
Moreover, 
we assume that the storage of the $n$ old gradients 
can be reduced to the 
storage of $n$ scalars, which is often the case in many machine learning models \cite{johnson2013accelerating}. 

A summary of 
the space requirements and convergence rates for
various stochastic ADMM algorithms is shown in Table~\ref{com_alg}. 
As can be seen, among those  with
variance reduction, 
the space requirements 
of SCAS-ADMM and SVRG-ADMM
are independent of the sample
size $n$. 
However, as discussed in the previous section, SVRG-ADMM has much faster convergence rates than SCAS-ADMM on both strongly convex and general convex problems.
%However, on strongly convex problems, SVRG-ADMM achieves a linear convergence rate while SCAS-ADMM only has sublinear %convergence.  On general convex problems, as discussed in the previous section, SVRG-ADMM has much faster sublinear rate than %SCAS-ADMM, especially when $n$ is large.

%%%%%%%%%%%%%%%%%%%%%%%%%%%%%%%%%%%%%%%%%%%%%%%%%%%%%%%%

\section{Experiments}
\label{sec:expt}

%%%%%%%%%%%%%%%%%%%%%%%%%%%%%%%%%%%%%%%%%%

\subsection{Graph-Guided Fused Lasso} 
\label{sec:expt1}

\begin{figure*}[ht]
\begin{center}
\subfigure{\includegraphics[width=.5\columnwidth, height=.325\columnwidth]{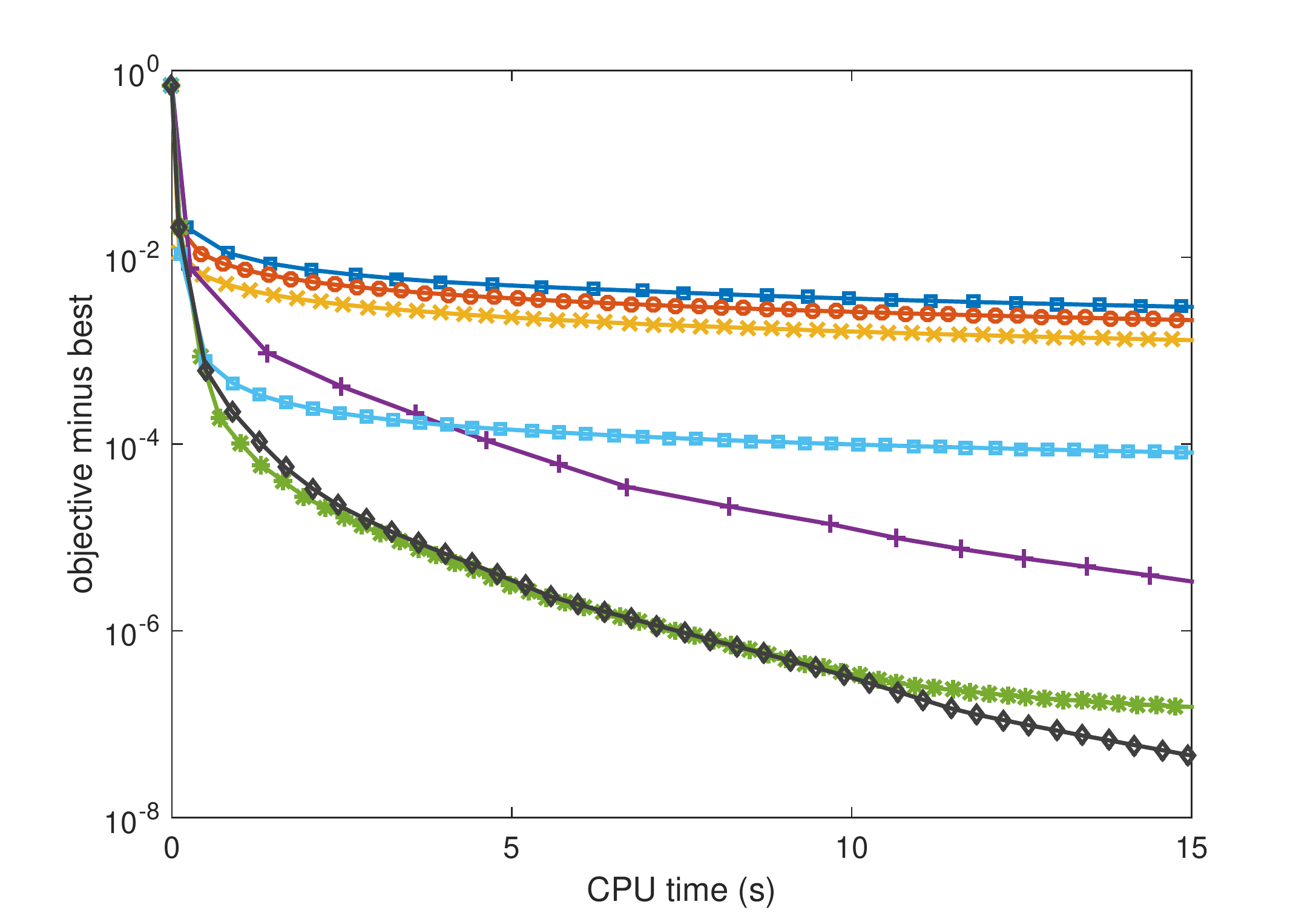}} 
\subfigure{\includegraphics[width=.5\columnwidth, height=.325\columnwidth]{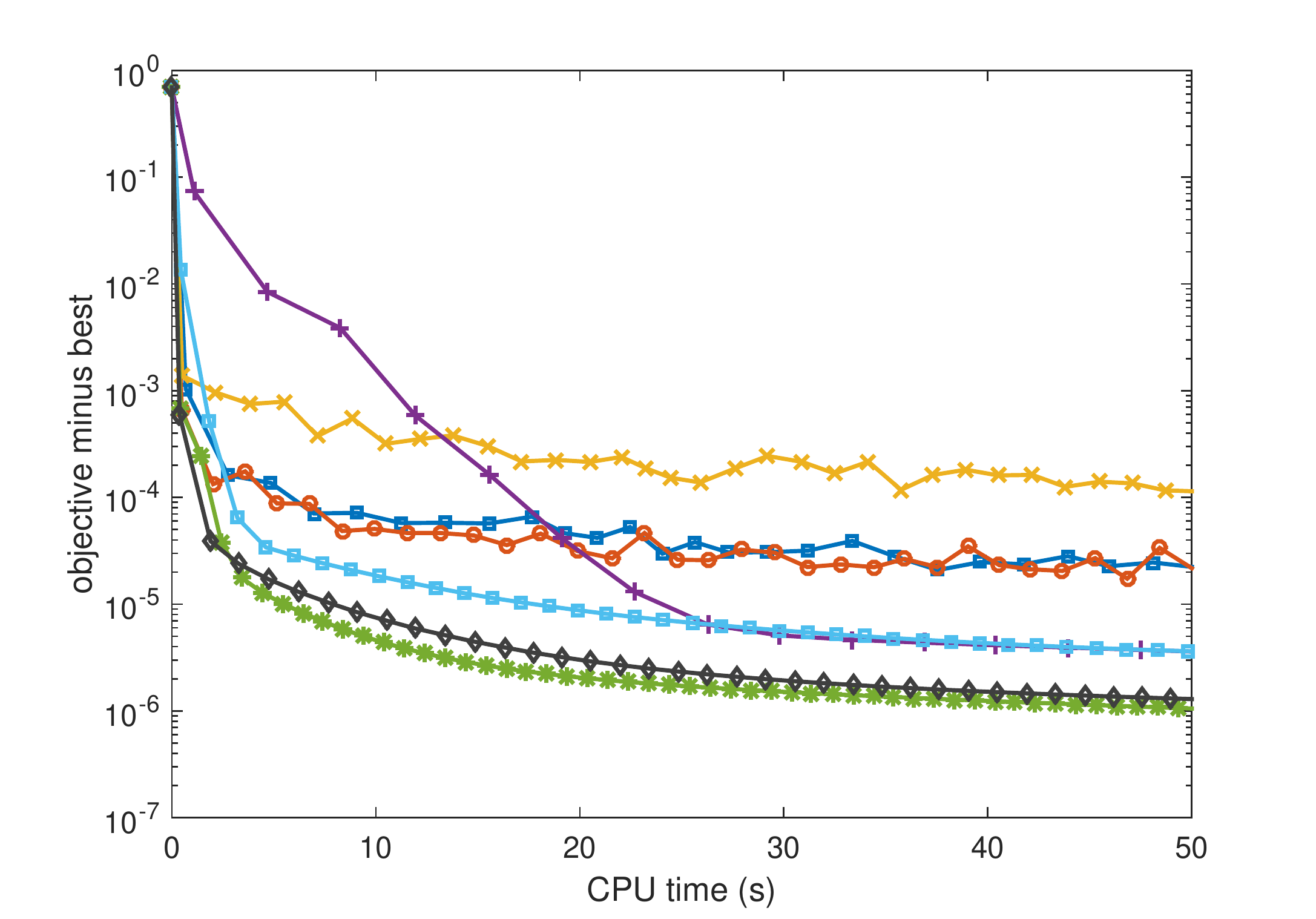}} 
\subfigure{\includegraphics[width=.5\columnwidth, height=.325\columnwidth]{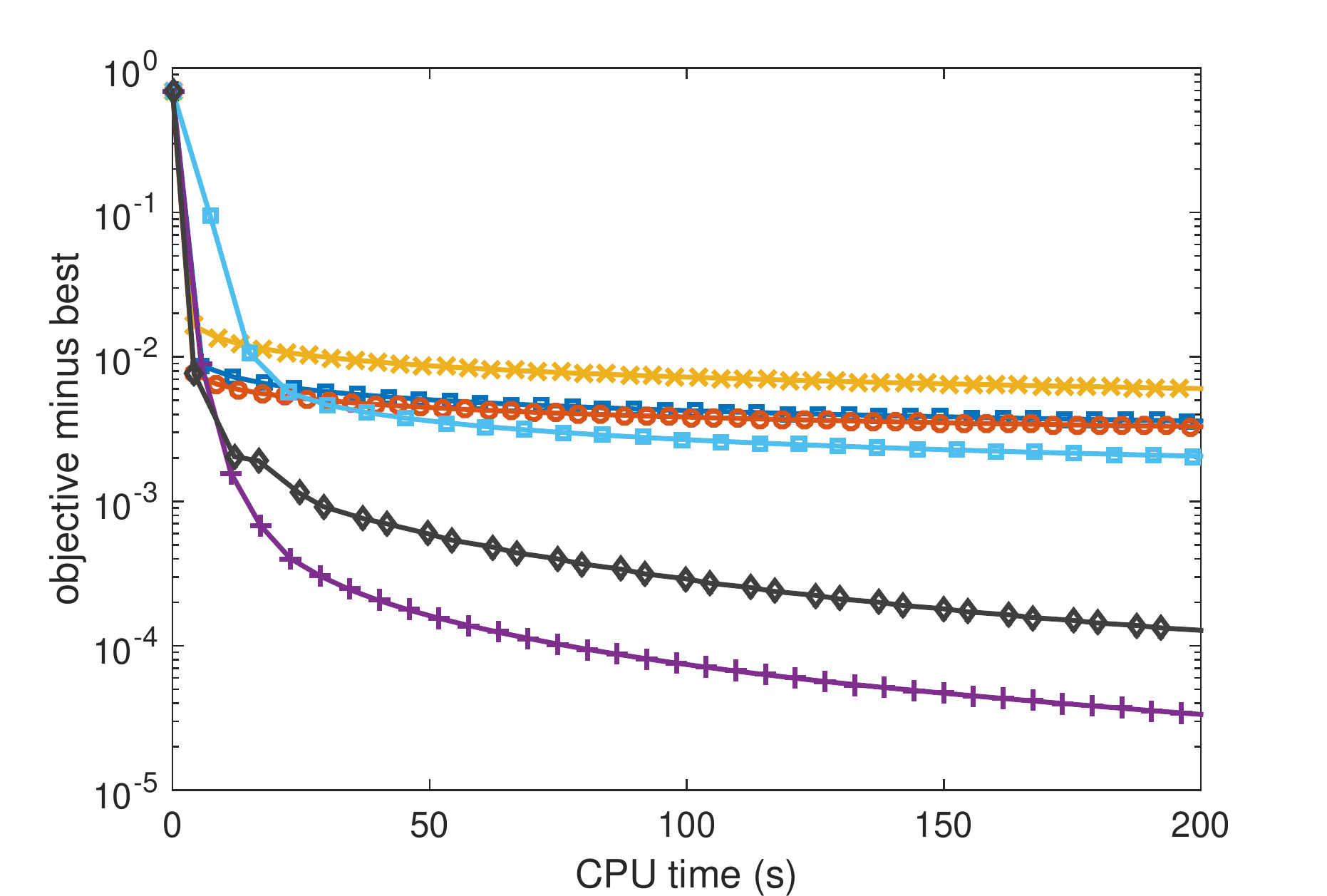}}
\subfigure{\includegraphics[width=.5\columnwidth, height=.325\columnwidth]{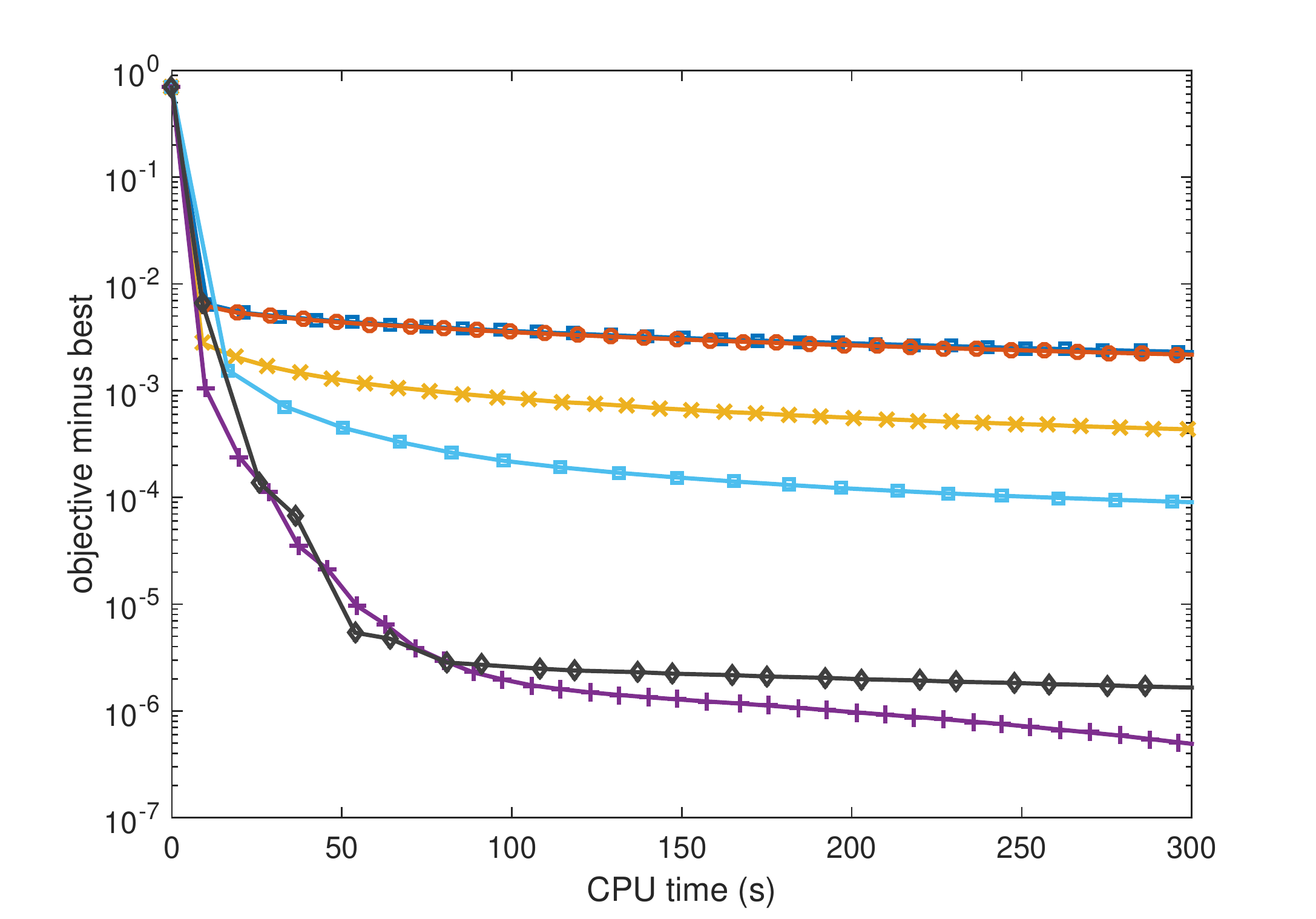}}
\setcounter{subfigure}{0}
\subfigure[{\em protein}]{\includegraphics[width=.5\columnwidth, height=.325\columnwidth]{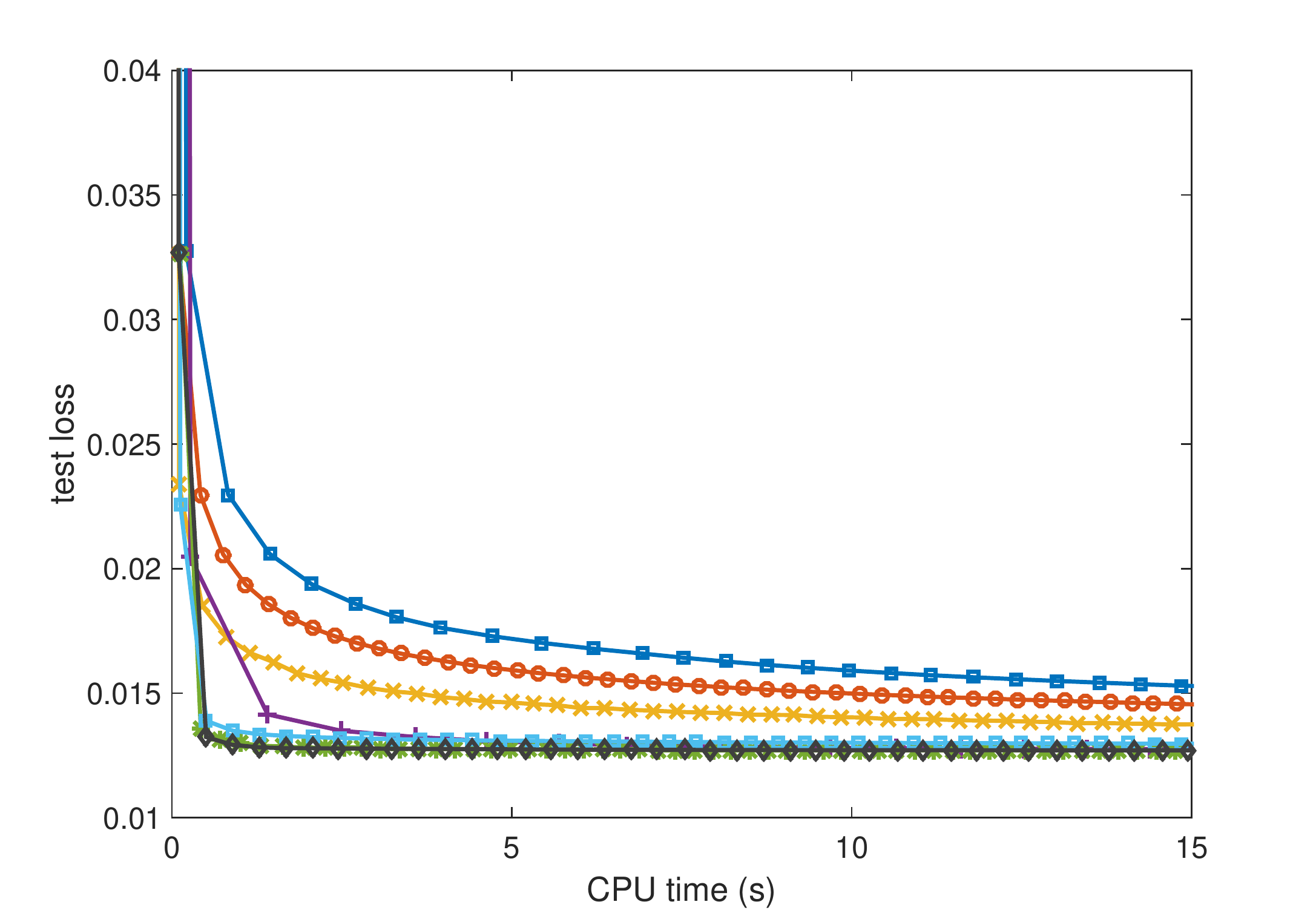}} 
\subfigure[{\em covertype}]{\includegraphics[width=.5\columnwidth, height=.325\columnwidth]{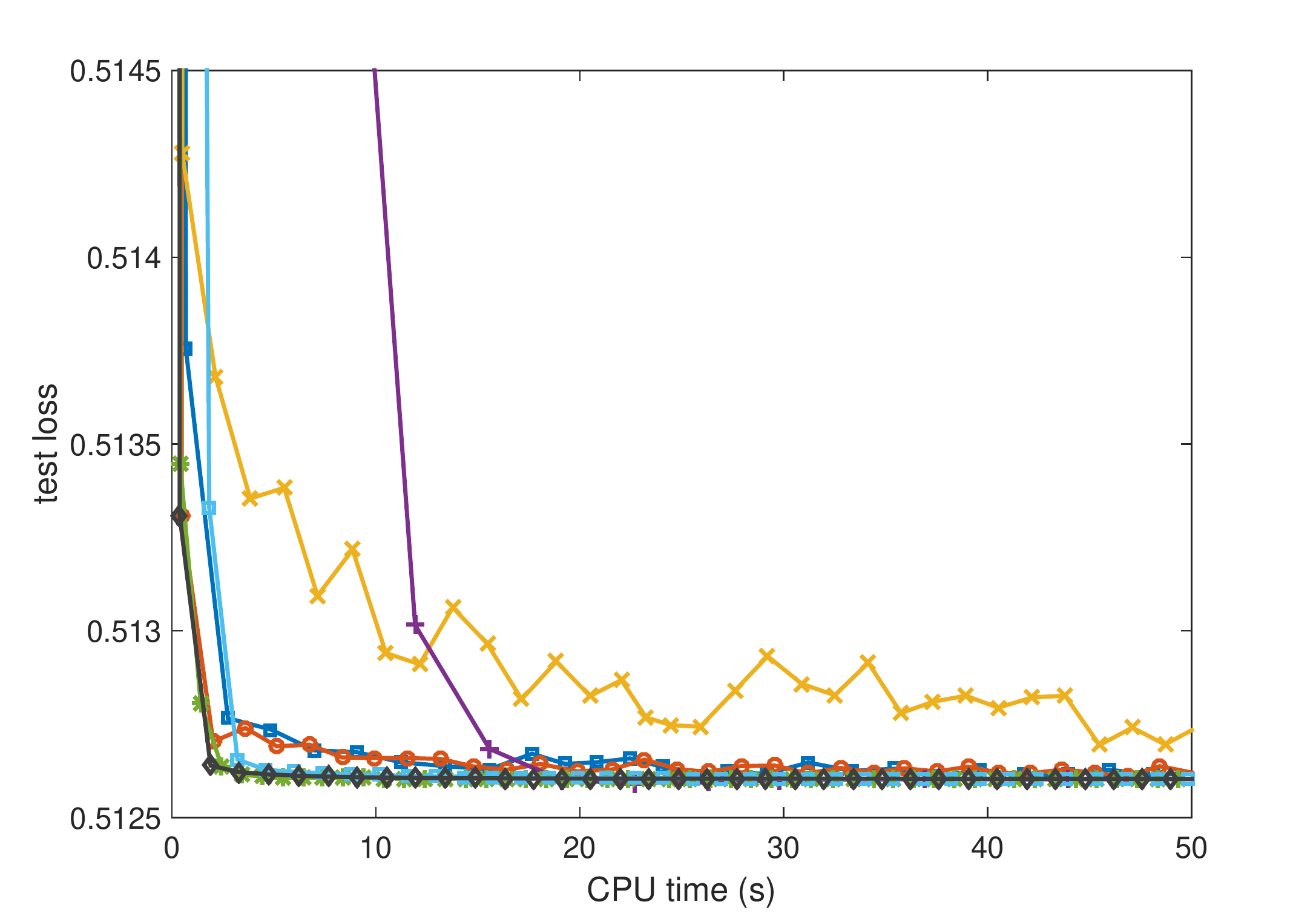}}
\subfigure[{\em mnist8m}]{\includegraphics[width=.5\columnwidth, height=.325\columnwidth]{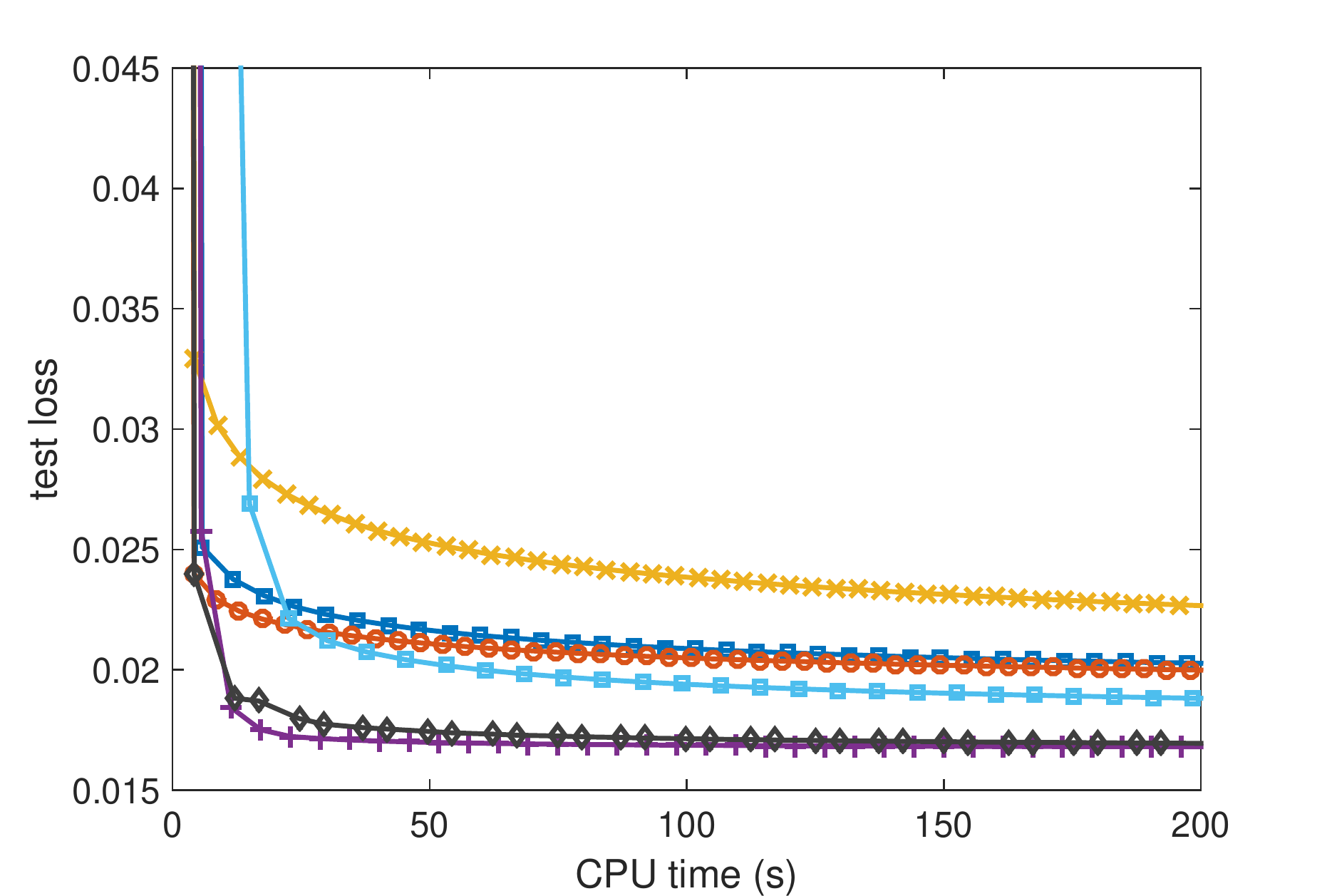}} 
\subfigure[{\em dna}]{\includegraphics[width=.5\columnwidth, height=.325\columnwidth]{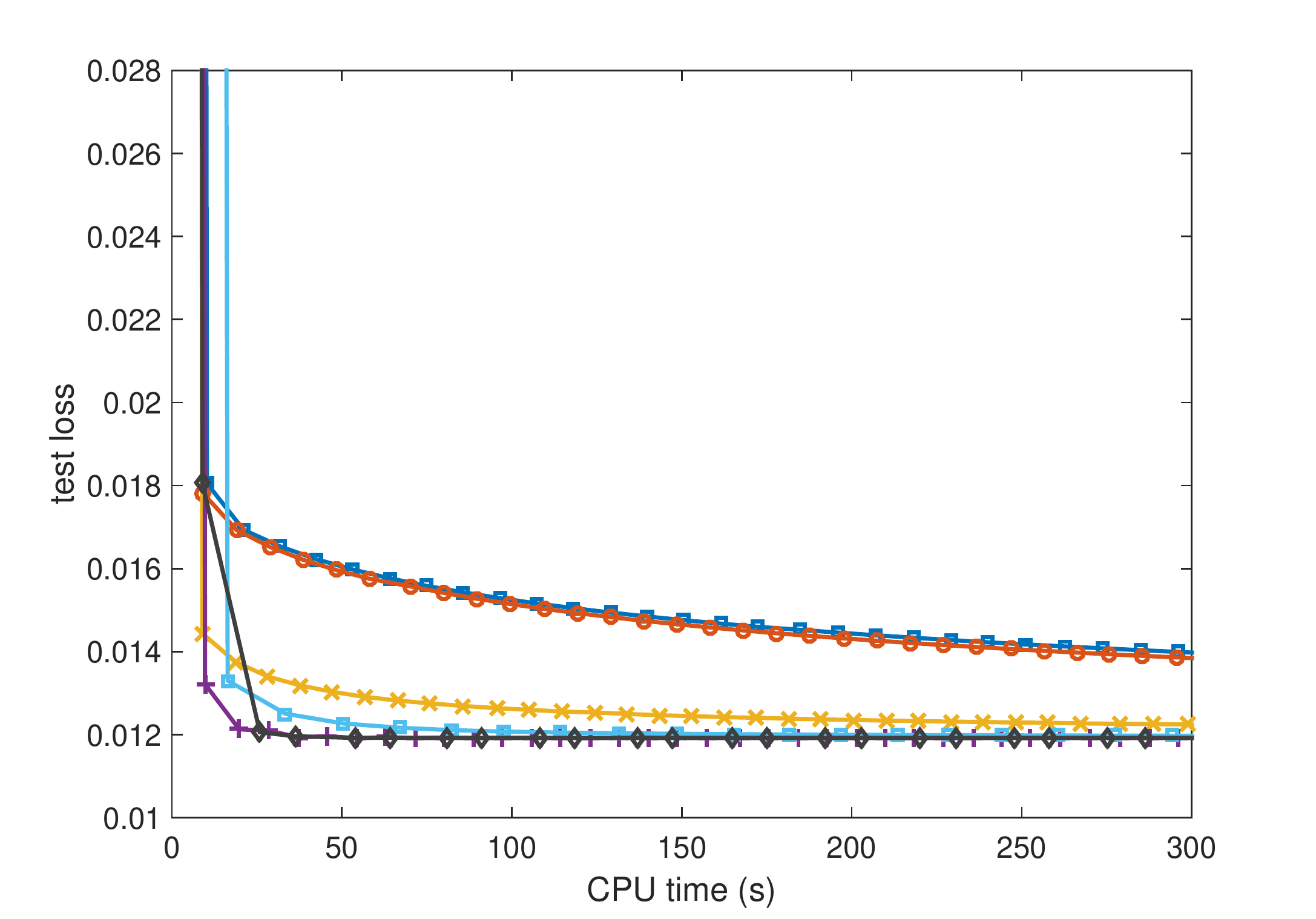}} 
\setcounter{subfigure}{0}
\includegraphics[width=6in,height=0.14in]{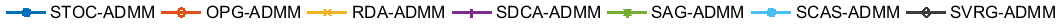}
\caption{Performance vs CPU time (in sec) on graph-guided fused lasso (Top: objective
value; Bottom: testing loss).}
\label{graph_lasso_time}
\end{center}
\end{figure*}

We perform
experiments on the 
generalized lasso model
$\sum_{i = 1}^n\ell_{i}(x) + \|Ax\|_1$, 
where $\ell_i$ is the logistic
loss on sample $i$, and $A$ is a matrix
encoding the feature sparsity pattern.  
Here, we use graph-guided fused lasso \cite{kim2009multivariate} 
and set $A = [G; I]$, 
where $G$ is the sparsity pattern of
the graph obtained by sparse inverse covariance estimation \cite{friedman2008sparse}.
For the ADMM formulation,
we introduce an additional variable $y$ and the constraint $Ax=y$. 
Experiments are performed on four
benchmark data
sets\footnote{Downloaded from
\url{http://www.csie.ntu.edu.tw/~cjlin/libsvmtools/datasets/},
\url{http://osmot.cs.cornell.edu/kddcup/datasets.html},
and \url{http://largescale.ml.tu-berlin.de/instructions/}.}
(Table~\ref{graph_lasso_data}).   
We use a mini-batch size of $b =100$
on {\em protein} and {\em covertype};
and $b = 500$
on {\em mnist8m} and {\em dna}.
Experiments are performed on a PC with Intel i7-3770 $3.4$GHz CPU and $32$GB RAM,

\begin{table}[ht]
\caption{Data sets for graph-guided fused lasso.}
\label{graph_lasso_data}
\begin{center}
\scalebox{1}{
\begin{tabular}{cccc}
\hline
& \#training & \#test & dimensionality \\ \hline
{\em protein}  &  72,876 & 72,875 & 74 \\
{\em covertype}   &  290,506 & 290,506 & 54         \\
{\em mnist8m}   &  1,404,756 & 351,189 & 784         \\
{\em dna}   &  2,400,000 & 600,000 & 800         \\
\hline
\end{tabular}
}
\end{center}
\end{table}

All methods listed in Table~\ref{com_alg} 
are compared
and in Matlab.
The proposed SVRG-ADMM uses the linearized update in (\ref{eq:svrg_iu}) and
$m = 2n/b$. 
For further speedup,
we simply use the last iterates 
in each stage 
($x_m, y_m, u_m$)
as $\tx_s,\ty_s,\tu_s$ in step~12 of
Algorithms~\ref{alg:admm_svrg} and \ref{alg:admm_svrg_general}.
Both SAG-ADMM and SVRG-ADMM are initialized by running OPG-ADMM for $n/b$ iterations.\footnote{This
extra CPU
time is counted towards the first stages of SAG-ADMM and SVRG-ADMM.}
For SVRG-ADMM, since the learning rate in (\ref{eq:svrg_iu}) is effectively $\eta/\gamma$, we set $\gamma = 1$ and only tune $\eta$.
All parameters are tuned as 
in \cite{zhong2014fast}.
Each stochastic algorithm is run on a small training subset 
for a few  
data passes  (or stages). 
The parameter setting with the smallest training objective 
is then chosen.
To ensure that the ADMM constraint is satisfied, 
we report the performance based on $(x_t, Ax_t)$. 
Results are averaged over five repetitions. 

Figure~\ref{graph_lasso_time} shows
the objective  values and testing losses 
versus 
CPU time.
SAG-ADMM cannot be run on {\em mnist8m} and {\em dna} because of its large memory requirement (storing the weights already takes 8.2GB for {\em mnist8m}, and 14.3GB for {\em dna}).
As can be seen, 
stochastic ADMM methods with variance reduction (SVRG-ADMM,
SAG-ADMM and SDCA-ADMM) have fast convergence,
while those that do not use variance reduction are much slower. SVRG-ADMM, SAG-ADMM
and SDCA-ADMM have comparable speeds, but SVRG-ADMM requires much less storage (see also
Table~\ref{com_alg}). 
On the medium-sized {\em protein} and {\em covertype},  
SCAS-ADMM has comparable  performance with the other stochastic ADMM variants using variance
reduction.
However, it becomes much slower on the larger {\em minist8m} and {\em dna}, which
is consistent with 
the analysis in 
Section~\ref{sec:discussion}.

%%%%%%%%%%%%%%%%%%%%%%%%%%%%%%%%%%%%%%%%%%%%%%%%%%%%%%%%%%%%%%%%%%%%%%%%%%%%%%

%%%%%%%%%%%%%%%%%%%%%%%%%%%%%%%%%%%%%%%%%%%
\subsection{Multitask Learning}

When there are a large number of outputs,
the much smaller space requirement
of SVRG-ADMM  is clearly 
advantageous.
In this section,  experiments are performed 
on an $1000$-class ImageNet data set  
\cite{russakovsky2014imagenet}. We use 1,281,167 images for training,
and $50,000$ images for 
testing.  
$4096$ features 
are extracted
from the last fully connected layer
of the convolutional net VGG-16 \cite{simonyan2014very}.
The multitask learning 
problem
is formulated as:
$\min_X \sum_{i = 1}^N\ell_{i}(X) + 
\lambda_1\|X\|_1 + \lambda_2\|X\|_{*}$,
where $X\in \R^{d\times N}$ is the parameter matrix,  
$N$ is the number of tasks,
$d$ is the feature dimensionality, 
$\ell_i$ is the multinomial logistic loss on the $i$th task, and $\|\cdot\|_*$ is the nuclear norm.
To solve this problem using ADMM, we introduce an additional variable $X'$ 
with the constraint $X'=X$. On setting
$A = [I; I]$,
the regularizer is then $g(AX) =
g([X; X']) =
\lambda_1\|X\|_1 + \lambda_2\|X'\|_{*}$. We set $\lambda_1 = 10^{-5}$, $\lambda_2 =
10^{-4}$, 
and use a mini-batch size $b = 500$.
SAG-ADMM requires 38.2TB for storing the weights, and SDCA-ADMM 9.6GB
for the dual variables, while SVRG-ADMM requires 62.5MB for storing $\tilde{x}$ and the full gradient.

\begin{figure}[ht]
\begin{center}
\subfigure[objective.]{\includegraphics[width=.49\columnwidth, height=.31\columnwidth]{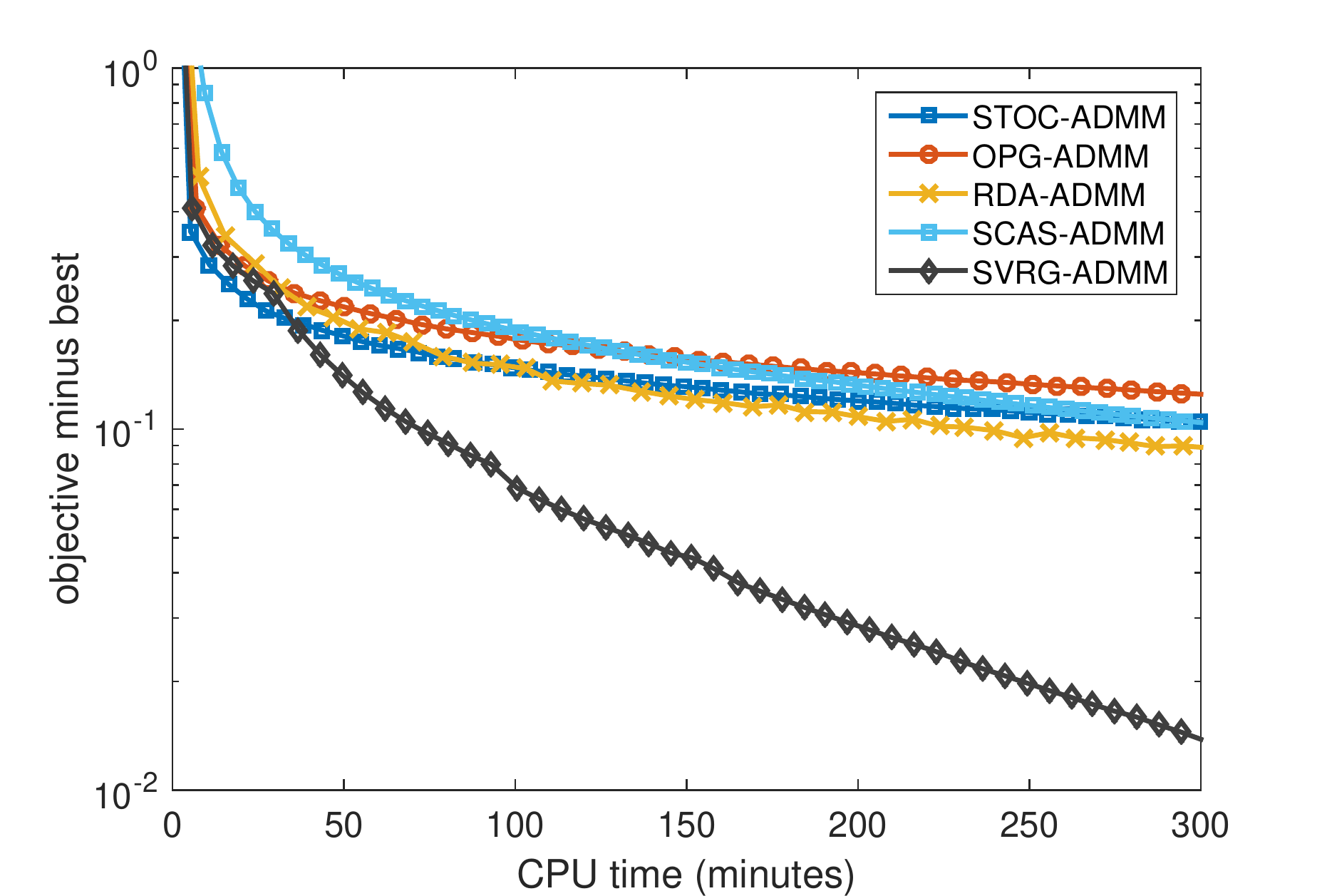}}
\subfigure[testing error (\%).]{\includegraphics[width=.49\columnwidth, height=.31\columnwidth]{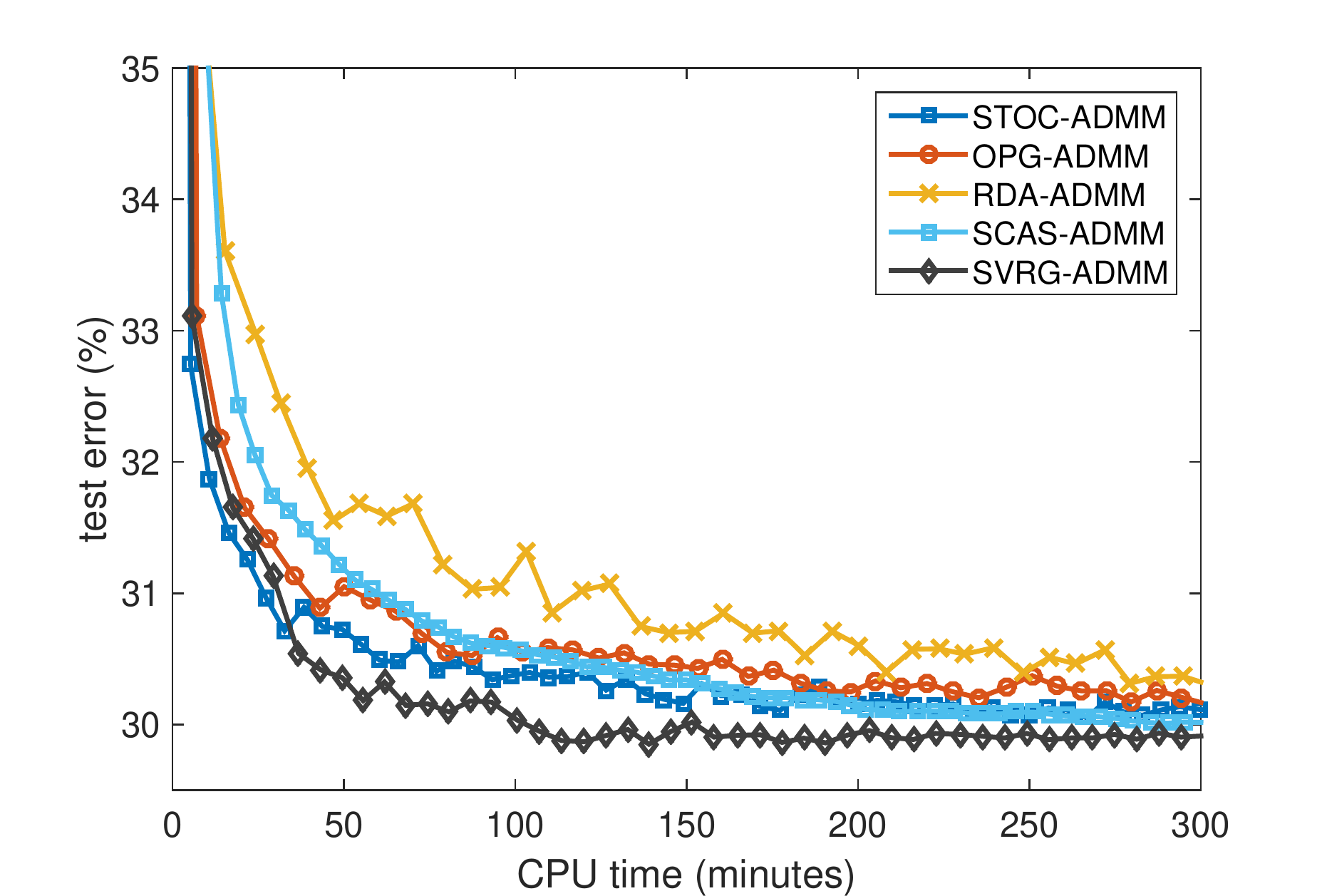}}
\end{center}
\caption{Performance vs CPU time (in min) on ImageNet.}
\label{sparse_low_rank_time}
\end{figure}

Figure~\ref{sparse_low_rank_time}
shows the objective value
and testing error
versus 
time.  
SVRG-ADMM converges rapidly to a good solution.
The other non-variance-reduced stochastic ADMM algorithms
are very aggressive initially, 
but quickly get much slower. SCAS-ADMM is again slow on this large data set.

%%%%%%%%%%%%%%%%%%%%%%%%%%%%%%%%%%%%%%%%%%

\subsection{Varying $\rho$}
\label{sec:vary}

Finally, we perform experiments on total-variation (TV) regression 
\cite{boyd2011distributed}
to demonstrate the effect of $\rho$.
Samples
$z_i$'s
are generated with 
i.i.d. 
components 
from the standard normal distribution. Each $z_i$ is then normalized to $\|z_i\| = 1$.
The parameter $x$ is generated according to 
{\small \url{http://www.stanford.edu/~boyd/papers/admm/}}. The output $o_i$ is obtained by adding standard
Gaussian noise
to $x^Tz_i$.  
Given $n$ samples $\{(z_1,o_1),\dots, (z_n,o_n)\}$, 
TV regression is formulated as:
$ \min_x \frac{1}{2n}\sum_{i=1}^n\|o_i - x^Tz_i\|^2
+ \lambda\|Ax\|_1$, where
$A_{ij} = 1$ if $i = j$; $-1$ if $j = i + 1$; and 0 otherwise.

\begin{figure}[h]
\begin{center}
\subfigure[objective.]{\includegraphics[width=.49\columnwidth, height=.31\columnwidth]{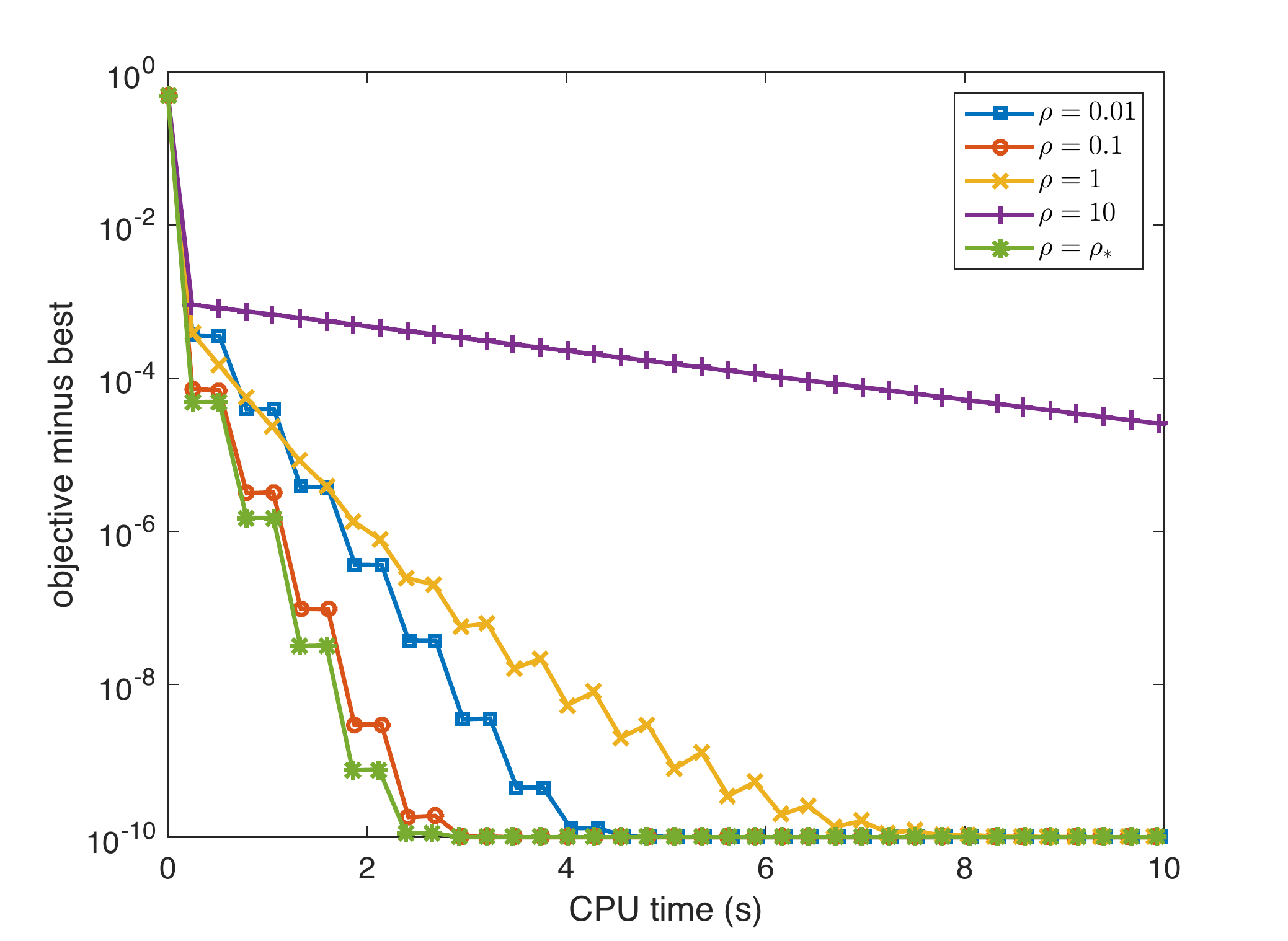}}
\subfigure[testing loss.]{\includegraphics[width=.49\columnwidth, height=.31\columnwidth]{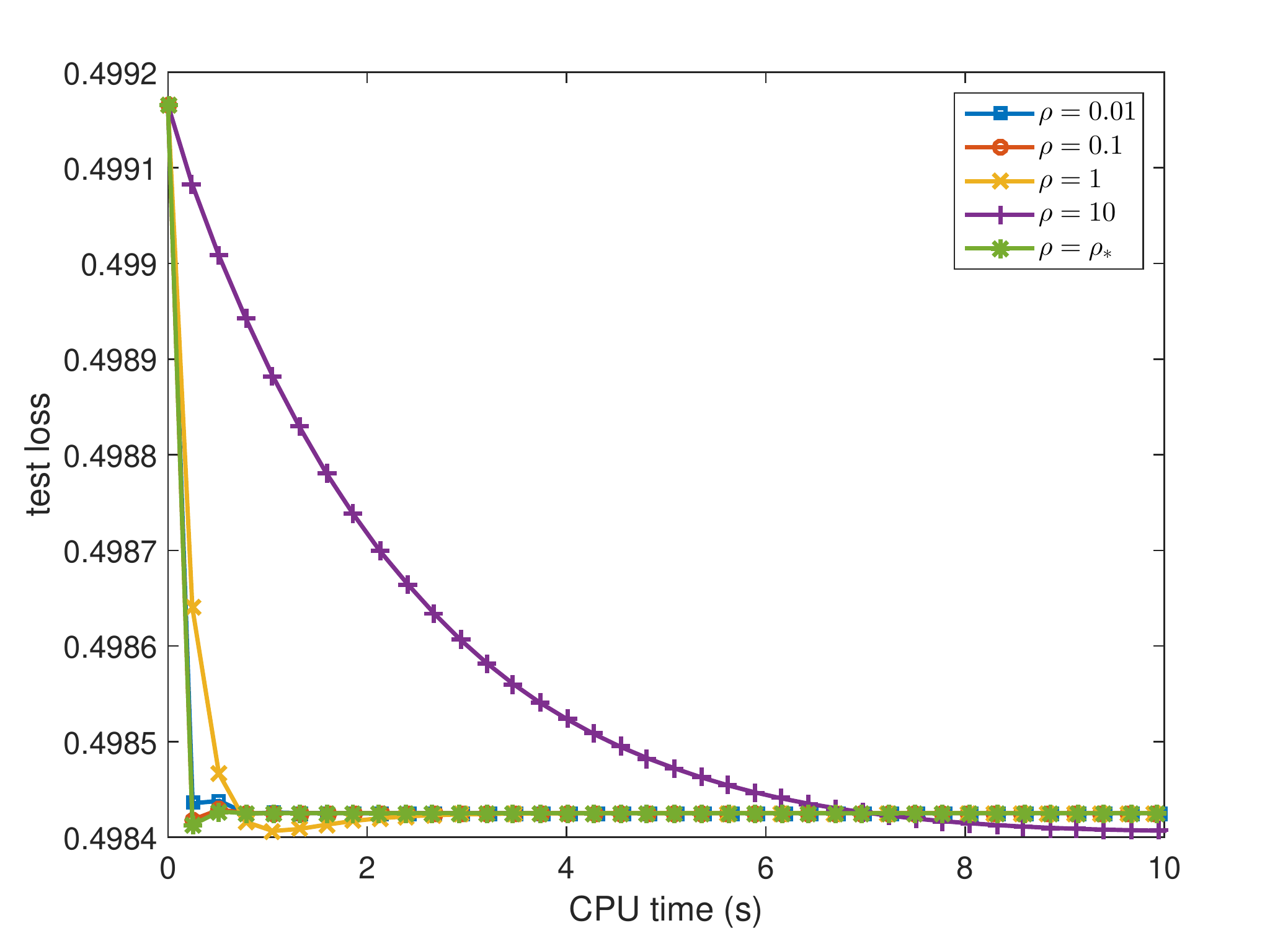}}
\end{center}
\caption{Performance of SVRG-ADMM at different $\rho$'s.}
\label{fig:tv}
\end{figure}

We set $n = 100,000, d = 500$,
$\lambda = 0.1/\sqrt{n}$,
and a mini-batch size $b = 100$.
Figure~\ref{fig:tv} shows the objective value and testing loss
versus CPU time,
with different 
$\rho$'s.
As can be seen, $\rho_*$ in Proposition~\ref{prop:kappa} outperforms the other choices of $\rho$.

\subsection{Nonconvex Graph-Guided Fused Lasso} 

\begin{figure}[h]
\begin{center}
\setcounter{subfigure}{0}
\subfigure[{\em a9a}.]{\includegraphics[width=.49\columnwidth, height=.335\columnwidth]{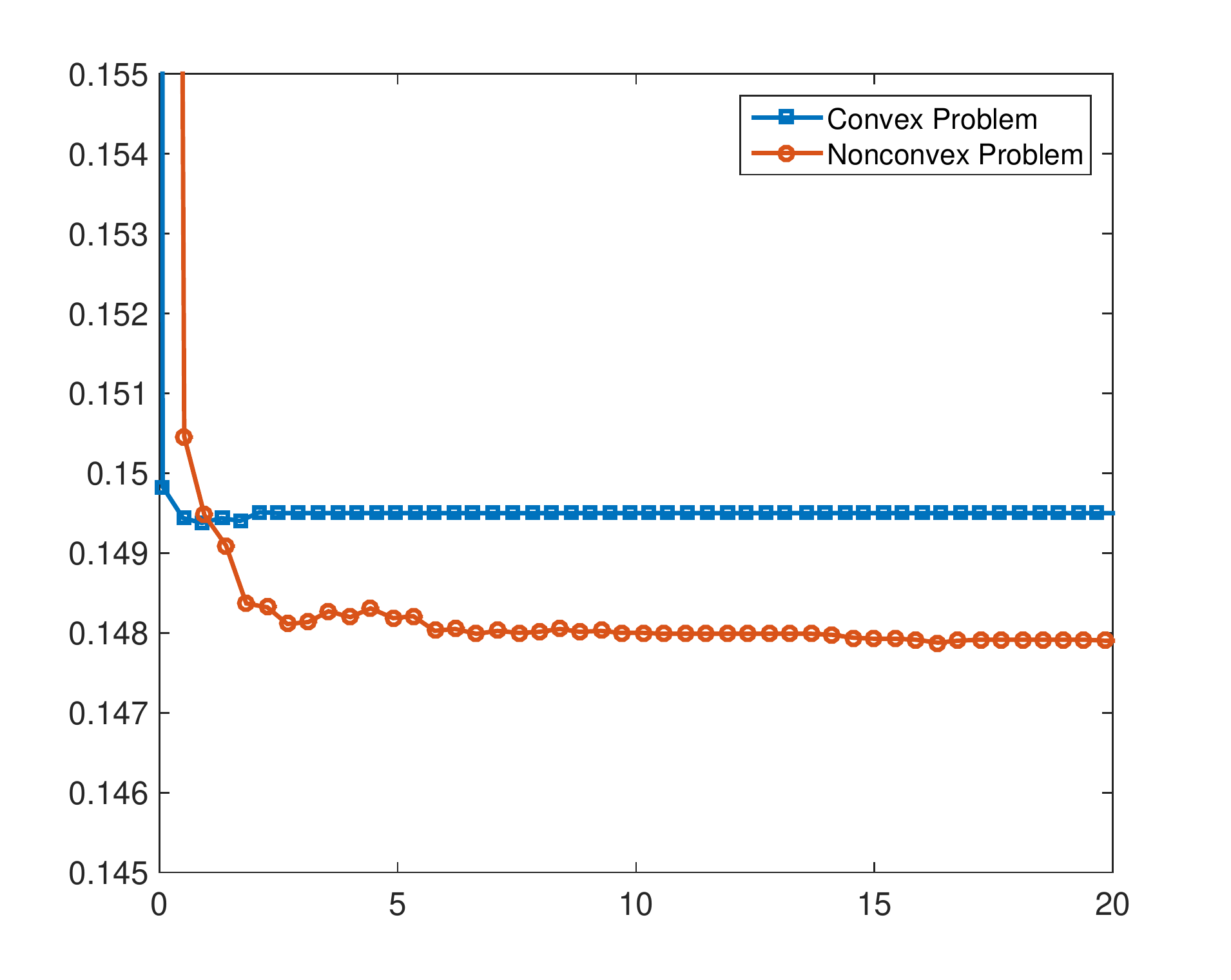}} 
\subfigure[{\em news20}.]{\includegraphics[width=.49\columnwidth, height=.335\columnwidth]{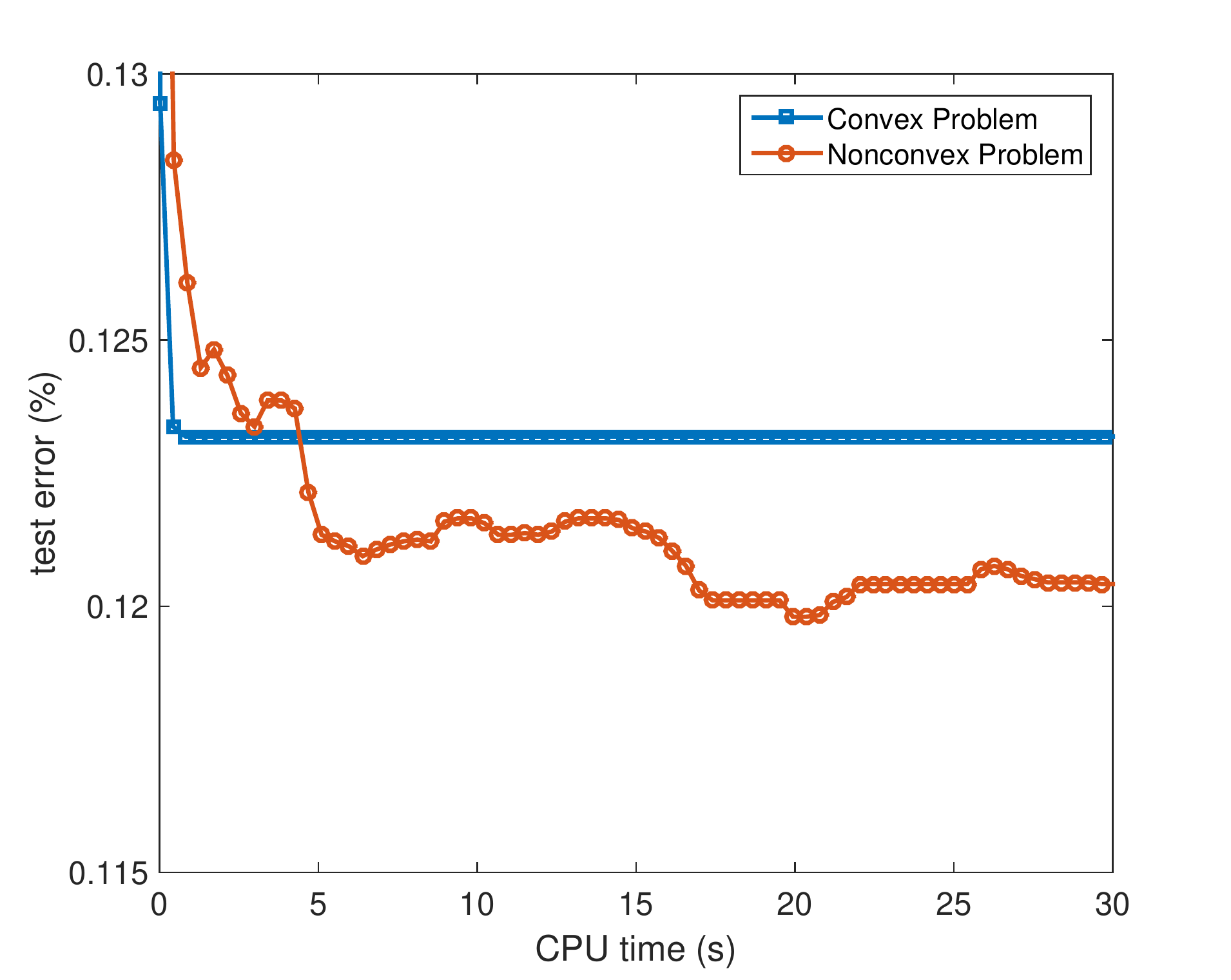}} 
\\
\subfigure[{\em protein}.]{\includegraphics[width=.49\columnwidth, height=.335\columnwidth]{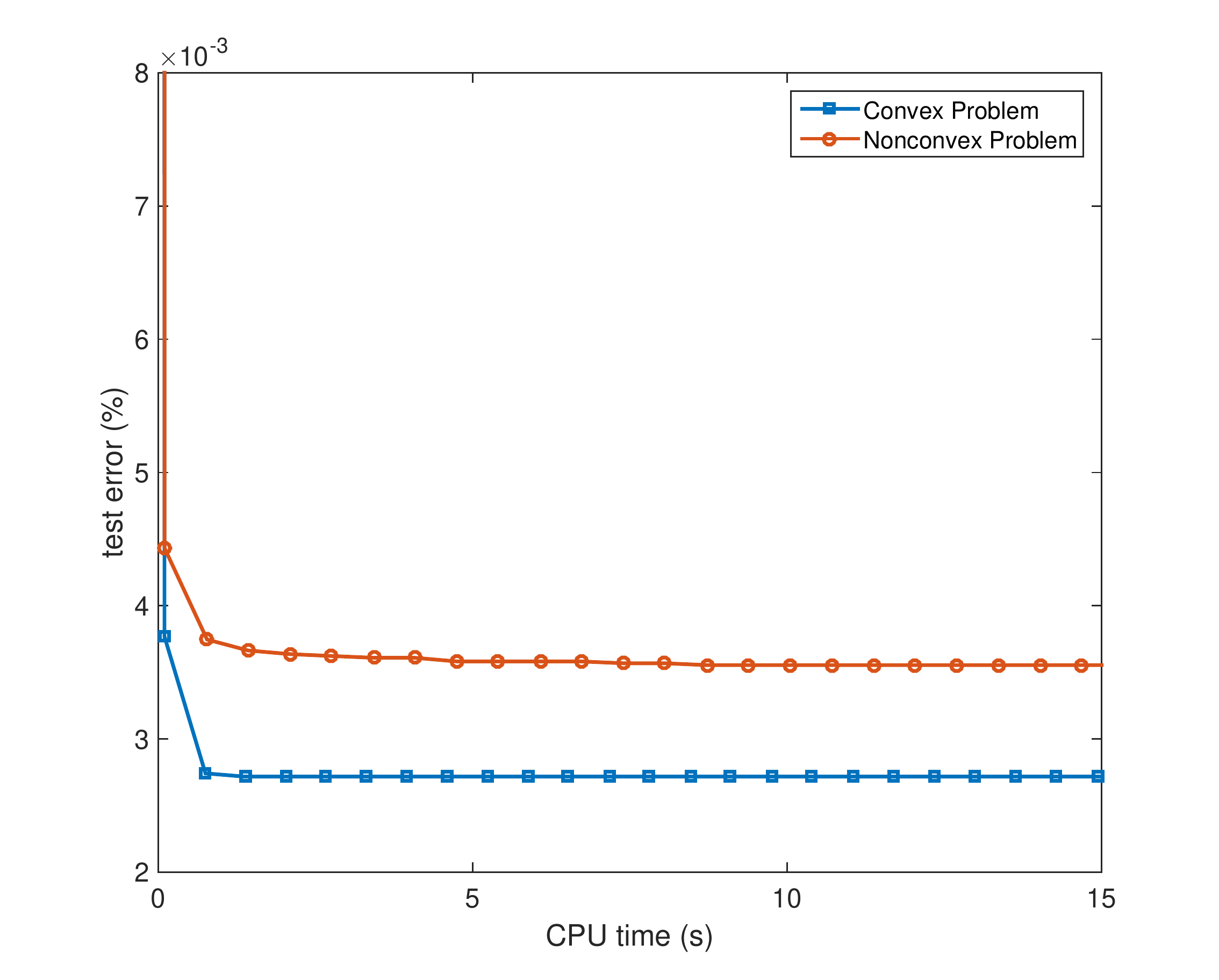}} 
\subfigure[{\em covertype}.]{\includegraphics[width=.49\columnwidth, height=.335\columnwidth]{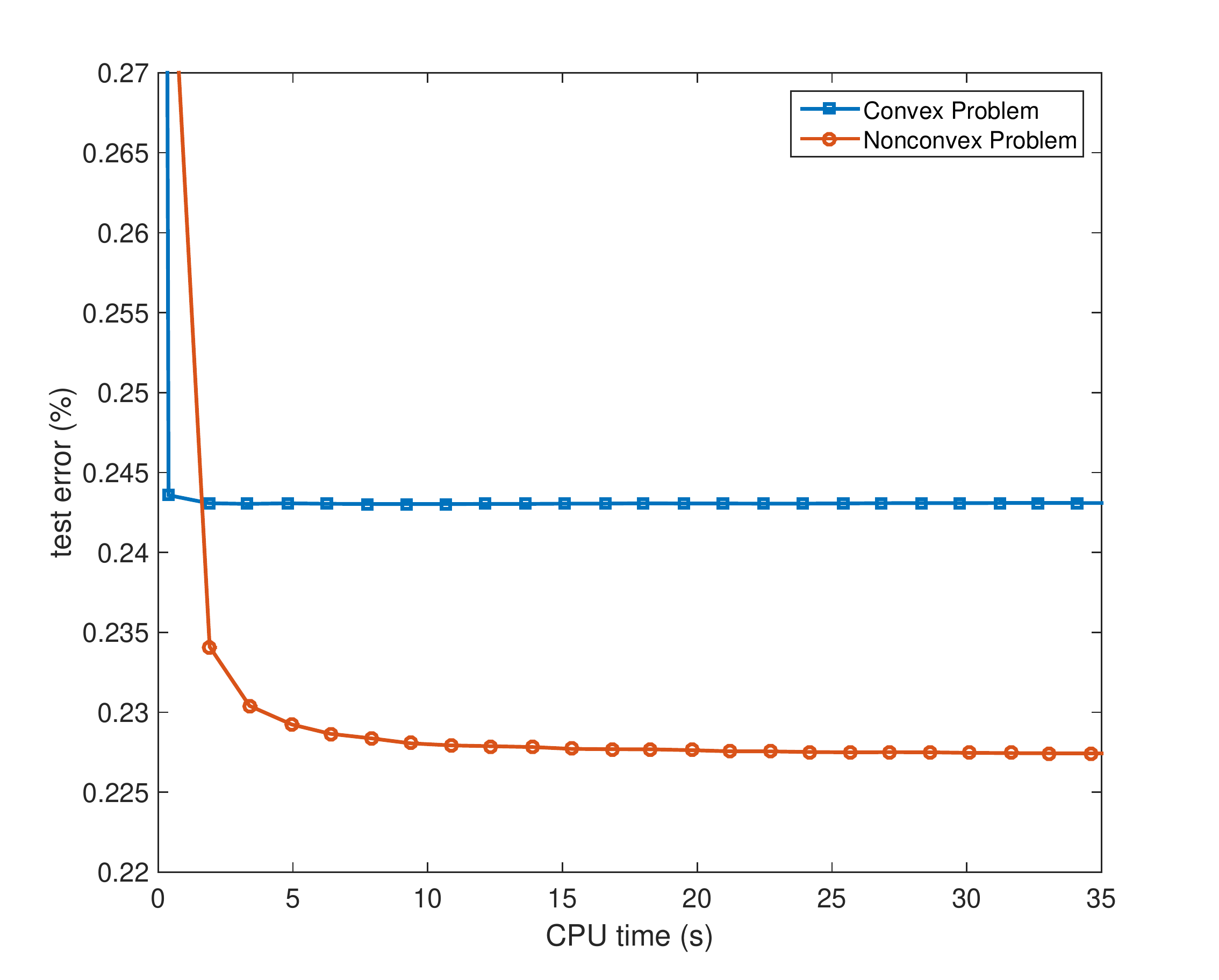}}
\setcounter{subfigure}{0}
\caption{Test error $(\%)$ vs CPU time (in sec) for the convex and nonconvex models.}
\label{nc_graph_lasso_time}
\end{center}
\end{figure}

In this section, we compare the performance of the convex and nonconvex graph-guided fused lasso models. The nonconvex graph-guided fused lasso model is given by
\[\frac{1}{n}\sum_{i = 1}^n\frac{1}{1 + \exp(o_iz_i^Tx)}  + \lambda\|Ax\|_1.\]
For the convex model, we simply replace the sigmod loss with the logistic loss. The data sets used are
summarized in Table~\ref{nc_graph_lasso_data}. Moreover, we use $\lambda = 10^{-4}$ for {\em a9a}, {\em
news20}, and $\lambda = 10^{-5}$ for {\em protein} and {\em covertype}. The test errors are shown in
Figure~\ref{nc_graph_lasso_time}. As can be seen, the nonconvex model obtains better results on the 
data sets
{\em a9a}, {\em news20} 
and {\em covertype}, while maintaining good convergence speed.

\begin{table}[ht]
\caption{Data sets for nonconvex graph-guided fused lasso.}
\label{nc_graph_lasso_data}
\begin{center}
%\scalebox{1}{
\begin{tabular}{cccc}
\hline
& \#training & \#test & dimensionality \\ \hline
{\em a9a}  &  32,561 & 16,281 & 123 \\
{\em news20}   &  12,995 & 3,247 & 100         \\
{\em protein}  &  72,876 & 72,875 & 74 \\
{\em covertype}   &  290,506 & 290,506 & 54         \\
\hline
\end{tabular}
%}
\end{center}
\end{table} 

\section{Conclusion}

This paper proposed a non-trivial integration of SVRG and ADMM.   
Its theoretical convergence rates for convex problems are as fast as
existing variance-reduced stochastic ADMM algorithms, 
but its storage requirement is much lower, even independent of the sample size. Besides,  we also show
the convergence rate of the proposed method on nonconvex problems.
Experimental results demonstrate its benefits over other stochastic ADMM methods and the benefits of
using a nonconvex model.

\bibliographystyle{IEEEtran}
\bibliography{admmsvrg}

% Generated by IEEEtran.bst, version: 1.13 (2008/09/30)
\begin{thebibliography}{10}
\providecommand{\url}[1]{#1}
\csname url@samestyle\endcsname
\providecommand{\newblock}{\relax}
\providecommand{\bibinfo}[2]{#2}
\providecommand{\BIBentrySTDinterwordspacing}{\spaceskip=0pt\relax}
\providecommand{\BIBentryALTinterwordstretchfactor}{4}
\providecommand{\BIBentryALTinterwordspacing}{\spaceskip=\fontdimen2\font plus
\BIBentryALTinterwordstretchfactor\fontdimen3\font minus
  \fontdimen4\font\relax}
\providecommand{\BIBforeignlanguage}[2]{{%
\expandafter\ifx\csname l@#1\endcsname\relax
\typeout{** WARNING: IEEEtran.bst: No hyphenation pattern has been}%
\typeout{** loaded for the language `#1'. Using the pattern for}%
\typeout{** the default language instead.}%
\else
\language=\csname l@#1\endcsname
\fi
#2}}
\providecommand{\BIBdecl}{\relax}
\BIBdecl

\bibitem{bottou-04}
L.~Bottou, ``Stochastic learning,'' in \emph{Advanced Lectures on Machine
  Learning}.\hskip 1em plus 0.5em minus 0.4em\relax Springer Verlag, 2004, pp.
  146--168.

\bibitem{parikh2014proximal}
N.~Parikh and S.~Boyd, ``Proximal algorithms,'' \emph{Foundations and Trends in
  Optimization}, vol.~1, no.~3, pp. 127--239, 2014.

\bibitem{defazio-14}
A.~Defazio, F.~Bach, and S.~Lacoste-Julien, ``{SAGA}: A fast incremental
  gradient method with support for non-strongly convex composite objectives,''
  in \emph{Advances in Neural Information Processing Systems}, 2014, pp.
  2116--2124.

\bibitem{johnson2013accelerating}
R.~Johnson and T.~Zhang, ``Accelerating stochastic gradient descent using
  predictive variance reduction,'' in \emph{Advances in Neural Information
  Processing Systems}, 2013, pp. 315--323.

\bibitem{roux2012stochastic}
N.~Roux, M.~Schmidt, and F.~Bach, ``A stochastic gradient method with an
  exponential convergence rate for finite training sets,'' in \emph{Advances in
  Neural Information Processing Systems}, 2012, pp. 2663--2671.

\bibitem{shalev2013stochastic}
S.~Shalev-Shwartz and T.~Zhang, ``Stochastic dual coordinate ascent methods for
  regularized loss,'' \emph{Journal of Machine Learning Research}, vol.~14,
  no.~1, pp. 567--599, 2013.

\bibitem{boyd2011distributed}
S.~Boyd, N.~Parikh, E.~Chu, B.~Peleato, and J.~Eckstein, ``Distributed
  optimization and statistical learning via the alternating direction method of
  multipliers,'' \emph{Foundations and Trends in Machine Learning}, vol.~3,
  no.~1, pp. 1--122, 2011.

\bibitem{ouyang2013stochastic}
H.~Ouyang, N.~He, L.~Tran, and A.~Gray, ``Stochastic alternating direction
  method of multipliers,'' in \emph{Proceedings of the 30th International
  Conference on Machine Learning}, 2013, pp. 80--88.

\bibitem{suzuki2013dual}
T.~Suzuki, ``Dual averaging and proximal gradient descent for online
  alternating direction multiplier method,'' in \emph{Proceedings of the 30th
  International Conference on Machine Learning}, 2013, pp. 392--400.

\bibitem{Wang2012}
H.~Wang and A.~Banerjee, ``Online alternating direction method,'' in
  \emph{Proceedings of the 29th International Conference on Machine Learning},
  2012, pp. 1119--1126.

\bibitem{zhong2014fast}
W.~Zhong and J.~Kwok, ``Fast stochastic alternating direction method of
  multipliers,'' in \emph{Proceedings of the 31st International Conference on
  Machine Learning}, 2014, pp. 46--54.

\bibitem{suzuki2014stochastic}
T.~Suzuki, ``Stochastic dual coordinate ascent with alternating direction
  method of multipliers,'' in \emph{Proceedings of the 31st International
  Conference on Machine Learning}, 2014, pp. 736--744.

\bibitem{zhao2015scalable}
S.~Y. Zhao, W.~J. Li, and Z.~H. Zhou, ``Scalable stochastic alternating
  direction method of multipliers,'' Tech. Rep. arXiv:1502.03529, 2015.

\bibitem{shen2014augmented}
Y.~Shen, Z.~Wen, and Y.~Zhang, ``Augmented lagrangian alternating direction
  method for matrix separation based on low-rank factorization,''
  \emph{Optimization Methods and Software}, vol.~29, no.~2, pp. 239--263, 2014.

\bibitem{liavas2015parallel}
A.~P. Liavas and N.~D. Sidiropoulos, ``Parallel algorithms for constrained
  tensor factorization via alternating direction method of multipliers,''
  \emph{IEEE Transactions on Signal Processing}, vol.~63, no.~20, pp.
  5450--5463, 2015.

\bibitem{jiang2016structured}
B.~Jiang, T.~Lin, S.~Ma, and S.~Zhang, ``Structured nonconvex and nonsmooth
  optimization: Algorithms and iteration complexity analysis,'' \emph{arXiv
  preprint arXiv:1605.02408}, 2016.

\bibitem{hong2016convergence}
M.~Hong, Z.~Q. Luo, and M.~Razaviyayn, ``Convergence analysis of alternating
  direction method of multipliers for a family of nonconvex problems,''
  \emph{SIAM Journal on Optimization}, vol.~26, no.~1, pp. 337--364, 2016.

\bibitem{li2015global}
G.~Li and T.~K. Pong, ``Global convergence of splitting methods for nonconvex
  composite optimization,'' \emph{SIAM Journal on Optimization}, vol.~25,
  no.~4, pp. 2434--2460, 2015.

\bibitem{wang2015convergence}
F.~Wang, W.~Cao, and Z.~Xu, ``Convergence of multi-block bregman admm for
  nonconvex composite problems,'' \emph{arXiv preprint arXiv:1505.03063}, 2015.

\bibitem{wang2015global}
Y.~Wang, W.~Yin, and J.~Zeng, ``Global convergence of admm in nonconvex
  nonsmooth optimization,'' \emph{arXiv preprint arXiv:1511.06324}, 2015.

\bibitem{nishihara2015general}
R.~Nishihara, L.~Lessard, B.~Recht, A.~Packard, and M.~I. Jordan, ``A general
  analysis of the convergence of {ADMM},'' in \emph{Proceedings of the 32nd
  International Conference on Machine Learning}, 2015, pp. 343--352.

\bibitem{deng2012global}
W.~Deng and W.~Yin, ``On the global and linear convergence of the generalized
  alternating direction method of multipliers,'' \emph{Journal of Scientific
  Computing}, pp. 1--28, 2015.

\bibitem{kim2009multivariate}
S.~Kim, K.~A. Sohn, and E.~P. Xing, ``A multivariate regression approach to
  association analysis of a quantitative trait network,''
  \emph{Bioinformatics}, vol.~25, no.~12, pp. i204--i212, 2009.

\bibitem{jacob2009group}
L.~Jacob, G.~Obozinski, and J.-P. Vert, ``Group lasso with overlap and graph
  lasso,'' in \emph{Proceedings of the 26th Annual International Conference on
  Machine Learning}, 2009, pp. 433--440.

\bibitem{ghadimi-14}
E.~Ghadimi, A.~Teixeira, I.~Shames, and M.~Johansson, ``Optimal parameter
  selection for the alternating direction method of multipliers ({ADMM}):
  Quadratic problems,'' \emph{IEEE Transactions on Automatic Control}, vol.~60,
  no.~3, pp. 644--658, 2015.

\bibitem{giselsson2014diagonal}
P.~Giselsson and S.~Boyd, ``Diagonal scaling in {D}ouglas-{R}achford splitting
  and {ADMM},'' in \emph{Proceedings of the 53rd IEEE Conference on Decision
  and Control}, 2014.

\bibitem{zhang2011unified}
X.~Zhang, M.~Burger, and S.~Osher, ``A unified primal-dual algorithm framework
  based on {B}regman iteration,'' \emph{Journal of Scientific Computing},
  vol.~46, no.~1, pp. 20--46, 2011.

\bibitem{golub2012matrix}
G.~Golub and C.~Van~Loan, \emph{Matrix Computations}.\hskip 1em plus 0.5em
  minus 0.4em\relax JHU Press, 2012.

\bibitem{he20121}
B.~He and X.~Yuan, ``On the ${O}(1/n)$ convergence rate of the
  {D}ouglas-{R}achford alternating direction method,'' \emph{SIAM Journal on
  Numerical Analysis}, vol.~50, no.~2, pp. 700--709, 2012.

\bibitem{xiao2014proximal}
L.~Xiao and T.~Zhang, ``A proximal stochastic gradient method with progressive
  variance reduction,'' \emph{SIAM Journal on Optimization}, vol.~24, no.~4,
  2014.

\bibitem{rockafellar2009variational}
R.~T. Rockafellar and R.~J. Wets, \emph{Variational analysis}.\hskip 1em plus
  0.5em minus 0.4em\relax Springer Science \& Business Media, 2009, vol. 317.

\bibitem{friedman2008sparse}
J.~Friedman, T.~Hastie, and R.~Tibshirani, ``Sparse inverse covariance
  estimation with the graphical lasso,'' \emph{Biostatistics}, vol.~9, no.~3,
  pp. 432--441, 2008.

\bibitem{russakovsky2014imagenet}
O.~Russakovsky, J.~Deng, H.~Su, J.~Krause, S.~Satheesh, S.~Ma, Z.~Huang,
  A.~Karpathy, A.~Khosla, M.~Bernstein, A.~C. Berg, and L.~F.-F., ``Imagenet
  large scale visual recognition challenge,'' \emph{International Journal of
  Computer Vision}, vol. 115, no.~3, pp. 211--252, 2015.

\bibitem{simonyan2014very}
K.~Simonyan and A.~Zisserman, ``Very deep convolutional networks for
  large-scale image recognition,'' Tech. Rep. arXiv:1409.1556, 2014.

\end{thebibliography}
%%%%%%%%%%%%%%%%%%%%%%%%%%%%%%%%%%%%%%%%%%%%%%%%%%%%%%%%%%%%%%%%%

\end{document}